\documentclass[runningheads]{llncs}
\usepackage{graphicx}

\usepackage{tikz}
\usepackage{comment}
\usepackage{amsmath,amssymb} 
\usepackage{color}

\usepackage{caption}
\usepackage{subcaption}
\usepackage{lipsum}
\usepackage{url}
\usepackage{multirow}
\usepackage{rotating}
\usepackage{tabularx}

\usepackage[accsupp]{axessibility}  

\begin{document}
\pagestyle{headings}
\mainmatter
\def\ECCVSubNumber{100}  

\title{Learning Neural Radiance Fields from Multi-View Geometry} 

\titlerunning{Learning Neural Radiance Fields from Multi-View Geometry}
%
\author{Marco Orsingher\inst{1,2} \and
Paolo Zani\inst{2} \and
Paolo Medici\inst{2} \and
Massimo Bertozzi\inst{1}}
\authorrunning{M. Orsingher et al.}
%
\institute{Università degli Studi di Parma \and
VisLab Srl (an Ambarella Inc company) \\
\email{marco.orsingher@unipr.it}}
\maketitle

\begin{abstract}

We present a framework, called \textbf{MVG-NeRF}, that combines classical \textbf{M}ulti-\textbf{V}iew \textbf{G}eometry algorithms and \textbf{Ne}ural \textbf{R}adiance \textbf{F}ields (NeRF) for image-based 3D reconstruction. NeRF has revolutionized the field of implicit 3D representations, mainly due to a differentiable volumetric rendering formulation that enables high-quality and geometry-aware novel view synthesis. However, the underlying geometry of the scene is not explicitly constrained during training, thus leading to noisy and incorrect results when extracting a mesh with marching cubes. To this end, we propose to leverage pixelwise depths and normals from a classical 3D reconstruction pipeline as geometric priors to guide NeRF optimization. Such priors are used as \textit{pseudo}-ground truth during training in order to improve the quality of the estimated underlying surface. Moreover, each pixel is weighted by a confidence value based on the forward-backward reprojection error for additional robustness. Experimental results on real-world data demonstrate the effectiveness of this approach in obtaining clean 3D meshes from images, while maintaining competitive performances in novel view synthesis. 

\end{abstract}

\section{Introduction}

\setcounter{footnote}{0}

3D reconstruction from a set of images is a longstanding problem in computer vision, with applications in robotics \cite{3drob1,3drob2}, virtual reality \cite{3daug1,3daug2}, autonomous driving \cite{3dself1,3dself2}, and many other fields. There are mainly two approaches in literature for inferring 3D geometry and appearance from multi-view images: (i) the classical geometry-based pipeline with Structure From Motion (SFM) \cite{sfm}, Multi-View Stereo (MVS) \cite{mvs} and Surface Reconstruction (SR) \cite{poisson,screened}; and (ii) modern deep learning methods \cite{nerf,mvsnet,atlas}, which can either replace single components in the classical pipeline or be trained end-to-end from images.

\begin{figure}[t]
     \centering
     \begin{subfigure}[b]{0.44\textwidth}
         \centering
         \includegraphics[width=\textwidth]{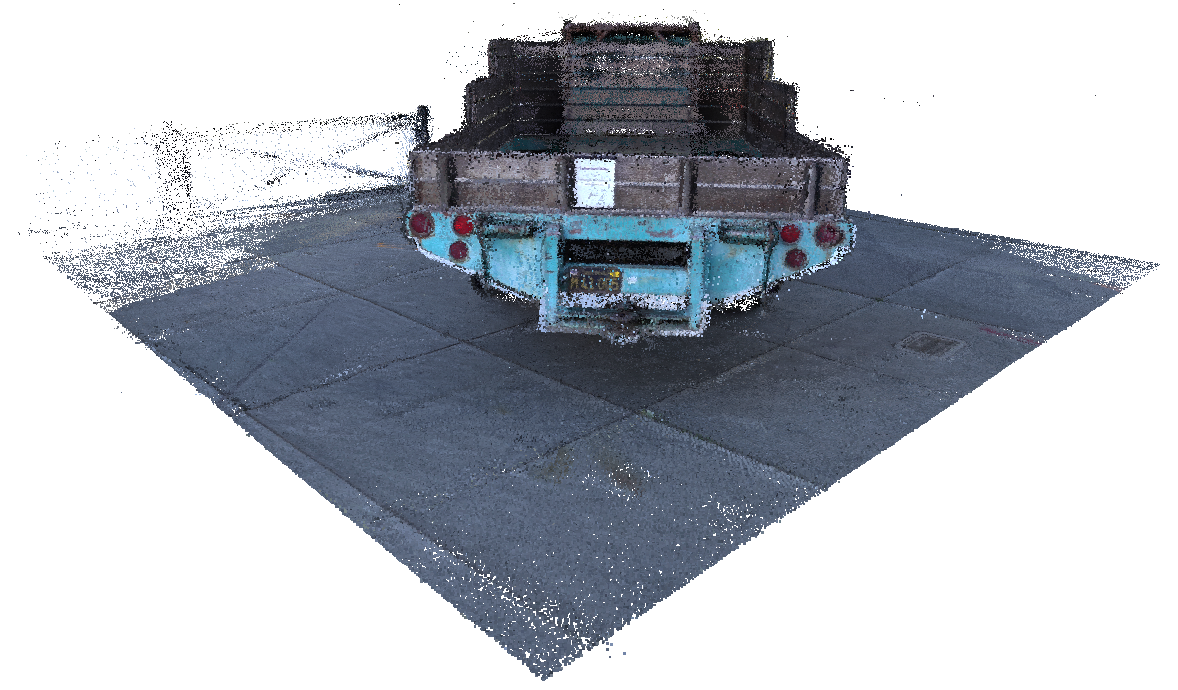}
         \caption{Input point cloud from MVS.}
         \label{fig:diff_pcd}
     \end{subfigure}
     \hfill
     \begin{subfigure}[b]{0.46\textwidth}
         \centering
         \includegraphics[width=\textwidth]{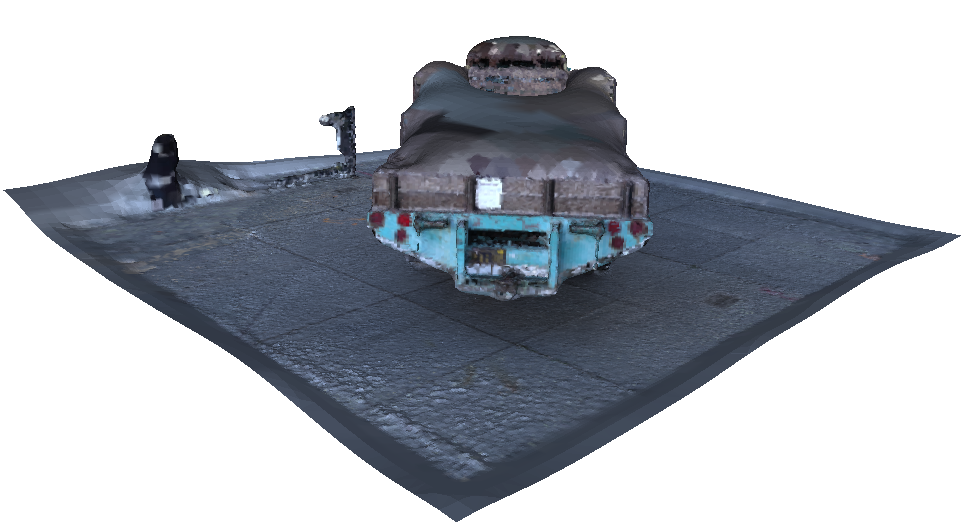}
         \caption{Mesh computed with SR.}
         \label{fig:diff_mesh}
     \end{subfigure}
     \begin{subfigure}[b]{0.5\textwidth}
         \centering
         \includegraphics[width=\textwidth]{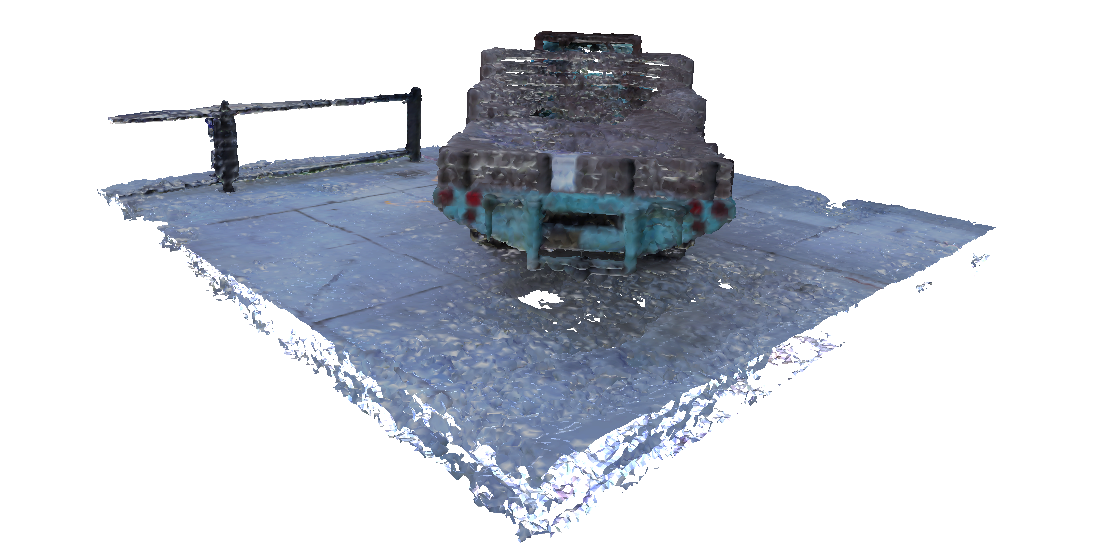}
         \caption{Mesh from NeRF.}
         \label{fig:diff_nerf}
     \end{subfigure}
     \hfill
     \begin{subfigure}[b]{0.48\textwidth}
         \centering
         \includegraphics[width=\textwidth]{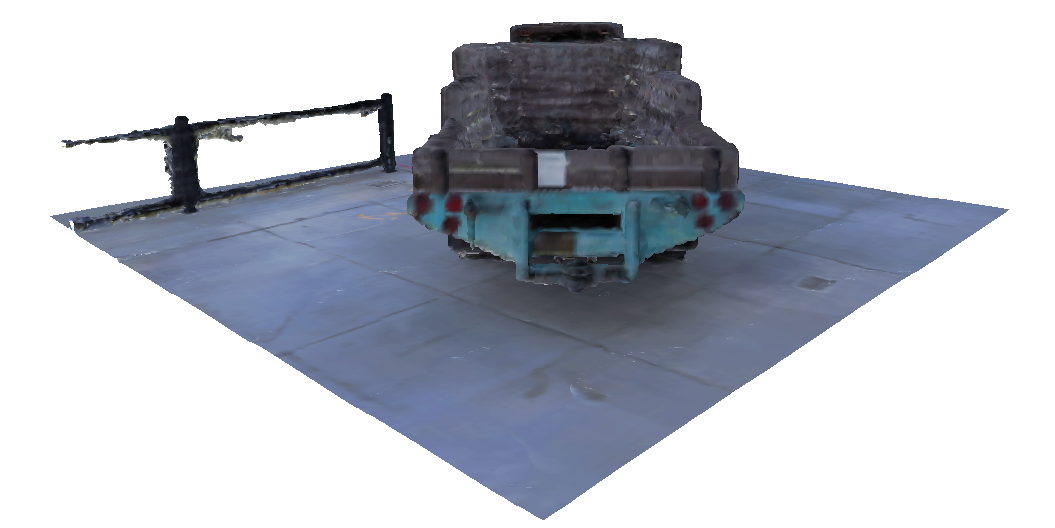}
         \caption{Mesh from our MVG-NeRF.}
         \label{fig:diff_ours}
     \end{subfigure}
        \caption{In complex structures such as the back of the truck, classical SR fails by hallucinating incorrect geometry, while the mesh from NeRF is very noisy. Our approach combines them to get a clean 3D model.}
        \label{fig:psr}
\end{figure}

A classical image-based 3D reconstruction pipeline takes a set of images as input and produces an \textit{explicit} representation of the scene, typically as a point cloud or as a mesh. Firstly, SFM \cite{sfm} computes camera poses and calibration parameters for each input image, as well as a set of sparse keypoints. Then, MVS \cite{mvs} estimates pixelwise depth and normal maps for each calibrated image. Finally, a point cloud can be obtained by backprojecting in 3D multi-view consistent hypotheses and a SR algorithm \cite{poisson,screened} is used to compute a mesh. Despite showing promising results in the last decades, this multi-stage approach suffers from errors that propagate between each stage, without any recovery mechanism. Moreover, MVS algorithms rely on photometric matching costs that are unreliable in textureless areas and non-Lambertian surfaces. Finally, while the \textit{discrete} representation of the scene as a point cloud is generally accurate, it is also locally sparse\footnote{This means that some areas of the scene are dense, while others are either empty or very sparse (usually where MVS fails).}, which makes it difficult for SR algorithms to recover the correct \textit{continuous} shape (see Fig. \ref{fig:diff_mesh}).

In recent years, there has been a growing interest in applying modern deep learning to 3D reconstruction. While some works proposed to augment the multi-stage pipeline with learning-based components \cite{pixsfm,mvsnet,shapeaspoints}, another research direction is to design \textit{implicit} 3D representations that encode shape and appearance in the weights of a neural network, such as a small MLP \cite{idr,dvr,nerf}. The concurrent development of differentiable surface \cite{idr,dvr} and volume \cite{nerf} rendering procedures allows end-to-end training from posed images, without 3D ground truth supervision. An \textit{explicit} representation of the scene as a mesh can then be extracted from the geometry field by querying the network with a dense grid of points and running the marching cubes algorithm \cite{marching}. On one hand, the main advantage of these approaches is that the 3D representation is \textit{continuous} by construction, as the network can be queried with any point in space. On the other hand, the underlying geometry is not explicitly constrained during training, which may lead to noisy and incorrect results, especially in real-world scenarios. A typical example is shown in Fig. \ref{fig:diff_nerf}.

We present a framework that combines neural implicit representations with classical multi-view geometry to produce clean 3D meshes from images, as shown in Fig. \ref{fig:diff_ours}. The key idea is to leverage pixelwise depths and normals from MVS as \textit{pseudo}-ground truth for constraining the underlying geometry of a Neural Radiance Field (NeRF) during training. Additionally, a confidence value is estimated for each pixel to softly activate this supervision only for rays with low reprojection error. Differently from recent works that include depth priors in NeRF \cite{dsnerf,nerfingmvs,ddpnerf}, we show that the joint optimization of normal vectors is crucial to improve the quality of the underlying surface. This is due to the fact that RGB and normals are complementary, meaning that normals can be estimated reliably in textureless regions where photometric consistency fails, while color supervision is effective in textured structures with ambiguous normals.

\section{Related Work}

The proposed approach is related to both classical 3D reconstruction (Sec. \ref{sec:cls}) and neural implicit 3D representations (Sec. \ref{sec:neu}), as well as the combination of these methods (Sec. \ref{sec:comb}). In this section, we briefly review the relevant literature in such fields.

\subsection{Classical 3D Reconstruction} 
\label{sec:cls}

The problem of recovering the 3D structure of a scene from a set of images has been studied for decades. SFM algorithms, which estimate extrinsic and intrinsic parameters for each image, can be broadly divided into global \cite{sfmglobal}, hierarchical \cite{sfmhierarchical} and incremental \cite{sfm} approaches. Given poses and calibration matrices, MVS then computes pixelwise depths and normals for each image. State-of-the-art MVS methods are mostly based on the PatchMatch idea to sample and propagate good hypotheses \cite{patchmatch,gipuma,mvs,acmm,acmp}, starting from a random initial solution. Finally, a 3D point cloud of the scene is given by the backprojection of multi-view consistent estimates and several SR methods have been developed to compute a continuous mesh \cite{poisson,screened}. However, SR often fails in intricate structures or relatively sparse regions, as shown in Fig. \ref{fig:psr}. While recent efforts focused on enhancing the classical pipeline with deep learning \cite{pixsfm,mvsnet,shapeaspoints}, purely geometry-based methods are still competitive on complex outdoor benchmarks \cite{eth3d,tnt}, do not require any 3D supervision for training and do not suffer from generalization issues. For this reason, we propose to exploit this readily available 3D information as \textit{pseudo}-ground truth for training a neural implicit 3D representation.

\subsection{Neural Implicit 3D Representations}
\label{sec:neu}

A recent line of research proposed to encode 3D geometry and appearance in small coordinate-based neural networks that can be queried with any point in space to produce the corresponding density \cite{nerf}, occupancy \cite{occupancy} or signed distance from the surface \cite{deepsdf}. While early approaches required 3D ground truth as supervision, differentiable surface \cite{idr,dvr} and volume \cite{nerf} rendering techniques have been introduced to enable self-supervised training from posed images alone. In particular, NeRF \cite{nerf} has shown impressive novel view synthesis results and opened several research directions to improve its training efficiency \cite{instant,regnerf}, memory consumption \cite{squeezenerf}, rendering speed \cite{instant,plenoctrees} and generalization \cite{geonerf,mvsnerf}. Moreover, the combination of surface and volume rendering \cite{unisurf,neus,volsdf} enables significant improvements in the scene geometry, at the cost of lower performances in novel view synthesis. Our idea is to keep the original formulation of NeRF for high-quality view synthesis and to use strong geometric priors to guide the optimization towards the correct underlying surface.

\begin{figure}[t]
    \centering
    \includegraphics[width=\textwidth]{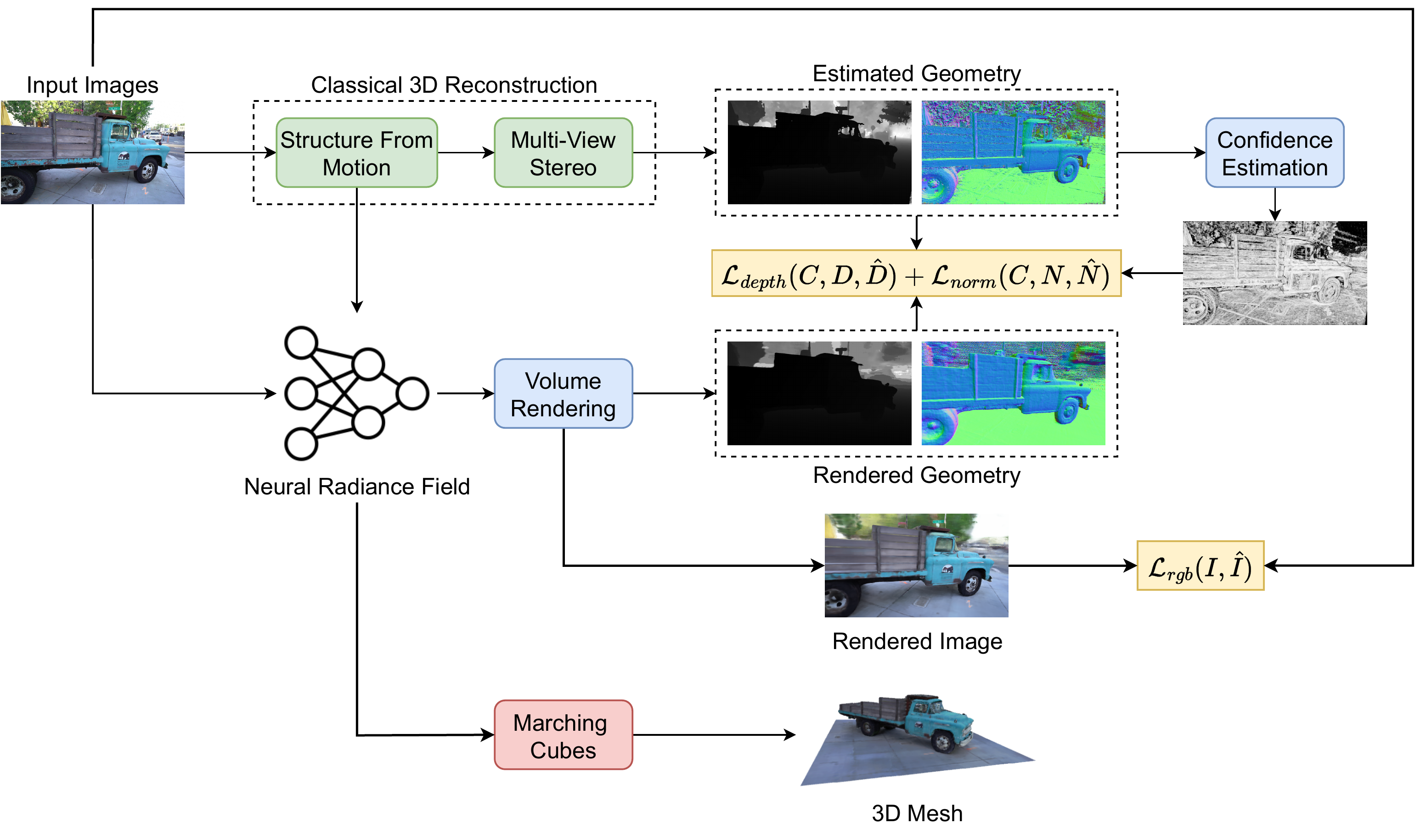}
    \caption{An overview of the approach. NeRF training (Sec. \ref{sec:nerf}) is guided by a confidence-aware (Sec. \ref{sec:conf}) geometric \textit{pseudo}-ground truth from classical 3D reconstruction. Pixelwise depths and normals are computed from NeRF with volume rendering (Sec. \ref{sec:geom}) and optimized with novel geometry losses (Sec. \ref{sec:loss}).}
    \label{fig:overview}
\end{figure}

\subsection{Geometric Priors for NeRF}
\label{sec:comb}

Several works have explored the possibility of introducing geometric priors while optimizing NeRF. DS-NeRF \cite{dsnerf} pioneered the idea of exploiting sparse keypoints from SFM and proposed a probabilistic depth supervision term in the loss function. Following this insight, NerfingMVS \cite{nerfingmvs} and DDP-NeRF \cite{ddpnerf} trained a monocular depth estimation and completion network, respectively, in order to have dense depth information for NeRF. These methods have shown to reduce both the number of views required for convergence and the overall training time. Inspired by these approaches, we instead propose to leverage pixelwise depths and normals from a classical 3D reconstruction pipeline as \textit{dense} and geometrically \textit{accurate} \textit{pseudo}-ground truth. MonoSDF \cite{monosdf} is a concurrent work that shows the importance of including normals as a geometric prior in the context of neural signed distance fields, but it relies on pretrained models for generating such priors.


\section{Method}

The proposed framework generates a 3D mesh of the scene from a set of raw images $\{I_i\}_{i=1}^N$, as shown in Fig. \ref{fig:overview}. Camera poses and calibration parameters $\{\mathbf{R}_i, \mathbf{t}_i, \mathbf{K}_i\}_{i=1}^N$ are estimated by a SFM algorithm and used as additional input for both NeRF and MVS. Then, pixelwise depths and normals $\{D_i, N_i\}_{i=1}^N$ are computed for each image with MVS and used to supervise NeRF training with confidence weights $\{C_i\}_{i=1}^N$. At the end of the training process, a 3D mesh is extracted from the density field with marching cubes \cite{marching}. 

After a short review of NeRF representation and color rendering procedure (Sec. \ref{sec:nerf}), we detail how to render geometric quantities (Sec. \ref{sec:geom}) and how to estimate the confidence of the \textit{pseudo}-ground truth (Sec. \ref{sec:conf}). Finally, loss functions are introduced and motivated in Sec. \ref{sec:loss}.

\subsection{Review of NeRF}
\label{sec:nerf}

\begin{figure}[t]
    \centering
    \includegraphics[width=\textwidth]{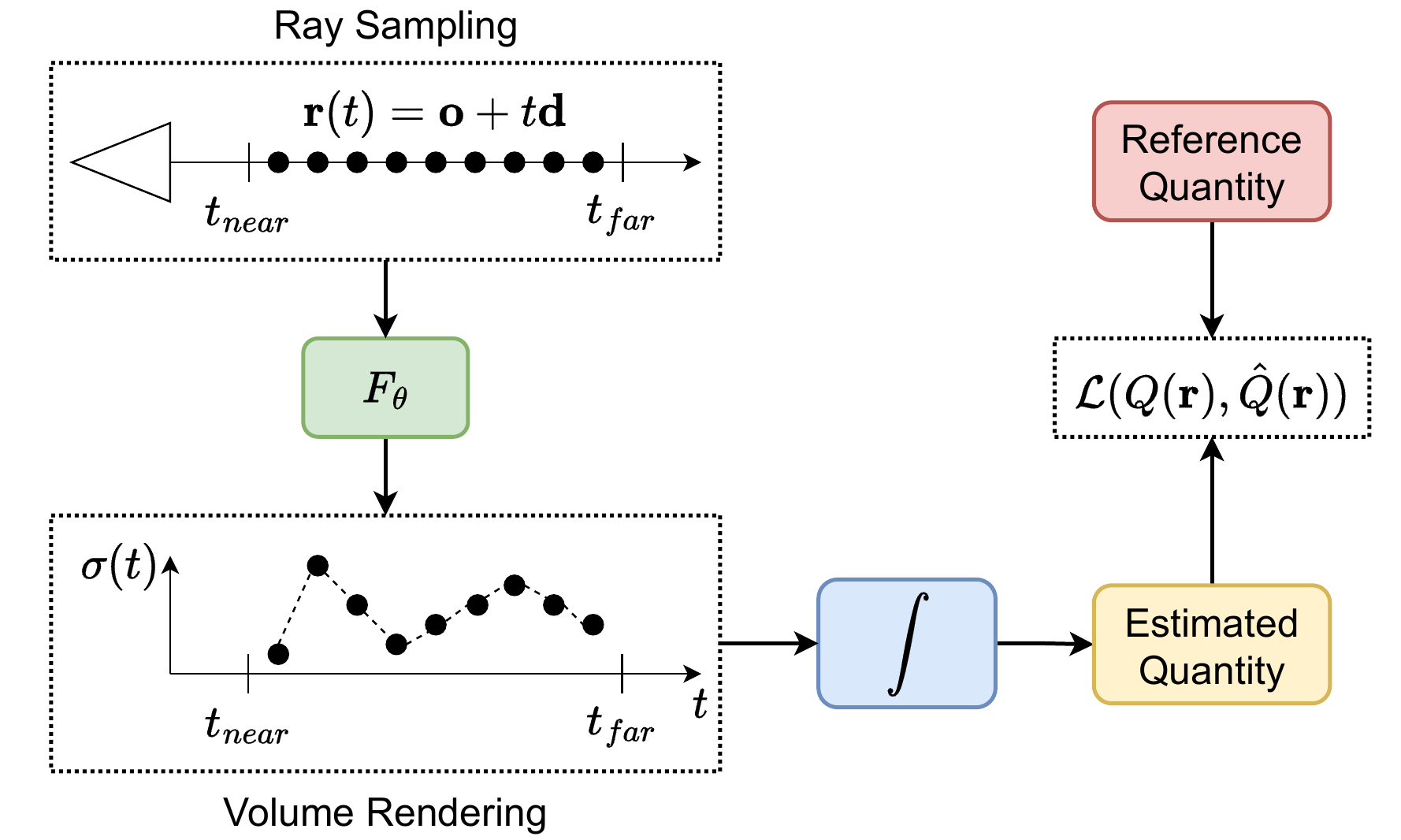}
    \caption{A visual explanation of differentiable volumetric rendering for arbitrary quantities $Q(\mathbf{r})$. A set of points is sampled along a ray $\mathbf{r}(t) = \mathbf{o} + t\mathbf{d}$, with $t \in [t_{near}, t_{far}]$. The network $F_\theta$ is queried at each point to produce the corresponding value, which is then integrated with weights based on the density $\sigma(t)$ to produce the final result $\hat{Q}(\mathbf{r})$. The goal is to minimize $\mathcal{L}(Q(\mathbf{r}),\hat{Q}(\mathbf{r}))$.}
    \label{fig:render}
\end{figure}

A Neural Radiance Field (NeRF) is an \textit{implicit} and \textit{continuous} representation of a 3D scene. Such field is implemented with a simple neural network $F_\theta$ that maps any point $\mathbf{x} \in \mathbb{R}^3$ in space and a viewing direction $\mathbf{d} \in \mathbb{S}^2$ to its corresponding density $\sigma(\mathbf{x}) \in \mathbb{R}^+$ and view-dependent color $\mathbf{c(x,d)} \in \mathbb{R}^3$:
\begin{equation}
    F_\theta : \mathbb{R}^3 \times \mathbb{S}^2 \rightarrow \mathbb{R}^3 \times \mathbb{R}^+
\end{equation}
This network is trained by minimizing an image reconstruction loss between the ground truth colors in the input images and the rendered colors from the radiance field. For each camera ray $\mathbf{r}(t) = \mathbf{o} + t\mathbf{d}$ with origin $\mathbf{o}$ and oriented as $\mathbf{d}$, the corresponding color $I(\mathbf{r})$ is computed by the following integral, bounded in the interval $[t_{near}, t_{far}]$:
\begin{equation}
\label{eq:render}
    \hat{I}(\mathbf{r}) = \int_{t_{near}}^{t_{far}} w(t) \cdot \mathbf{c}(t) dt
\end{equation}
with volumetric integration weights:
\begin{equation}
    w(t) = \text{exp}\left(-\int_{t_{near}}^t \sigma(s) ds \right) \cdot \sigma(t)
\end{equation}
In pratice, these integrals are approximated by numerical quadrature with a discrete set of samples along each ray. More details can be found in \cite{nerf} and a visual explanation of the rendering procedure is provided in Fig. \ref{fig:render}.

NeRF allows to (i) render novel views by sampling points along a ray through each pixel and integrating those samples with volume rendering; (ii) extract a 3D mesh of the scene by running the marching cubes algorithm \cite{marching} on the density field. However, the main issue is that NeRF is explicitly optimized only to produce RGB renderings and the underlying geometry is never constrained during training. This leads to the well-known phenomenon of shape-radiance ambiguity \cite{nerfplusplus}, which means that NeRF can hallucinate incorrect geometries as long as they explain the input views in terms of the image reconstruction loss.

\subsection{Rendering Geometry from NeRF}
\label{sec:geom}

The color rendering equation can be modified to render volumetrically any other quantity, including the underlying geometry that is learned during training. The expected depth along a ray can be computed as follows:
\begin{equation}
\label{eq:depthrender}
    \hat{D}(\mathbf{r}) = \int_{t_{near}}^{t_{far}} w(t) \cdot t dt
\end{equation}
Moreover, the unit normal vector $\mathbf{n(x)}$ at point $\mathbf{x} \in \mathbb{R}^3$ is given by the gradient of the density field at $\mathbf{x}$:
\begin{equation}
    \mathbf{n(x)} = \frac{\nabla_\mathbf{x}F_\theta(\mathbf{x})}{||\nabla_\mathbf{x}F_\theta(\mathbf{x})||}
\end{equation}
Such gradient could be directly computed from the automatic differentiation engine of deep learning frameworks. However, in our experiments, a simple central difference approximation has proven to be more accurate and efficient. Let $\Delta h$ be a small step size:
\begin{equation}
    \nabla_\mathbf{x}F_\theta(\mathbf{x}) \approx \frac{F_\theta(\mathbf{x} + \Delta h) - F_\theta(\mathbf{x} - \Delta h)}{\Delta h}
\end{equation}
Given the unit normal vectors for each sample along a ray, the final normal can be rendered as shown in Fig. \ref{fig:render}:
\begin{equation}
\label{eq:normalrender}
    \hat{N}(\mathbf{r}) = \int_{t_{near}}^{t_{far}} w(t) \cdot \mathbf{n}(t) dt
\end{equation}

\subsection{Confidence Estimation}
\label{sec:conf}

\begin{figure}
     \centering
     \includegraphics[width=0.75\textwidth]{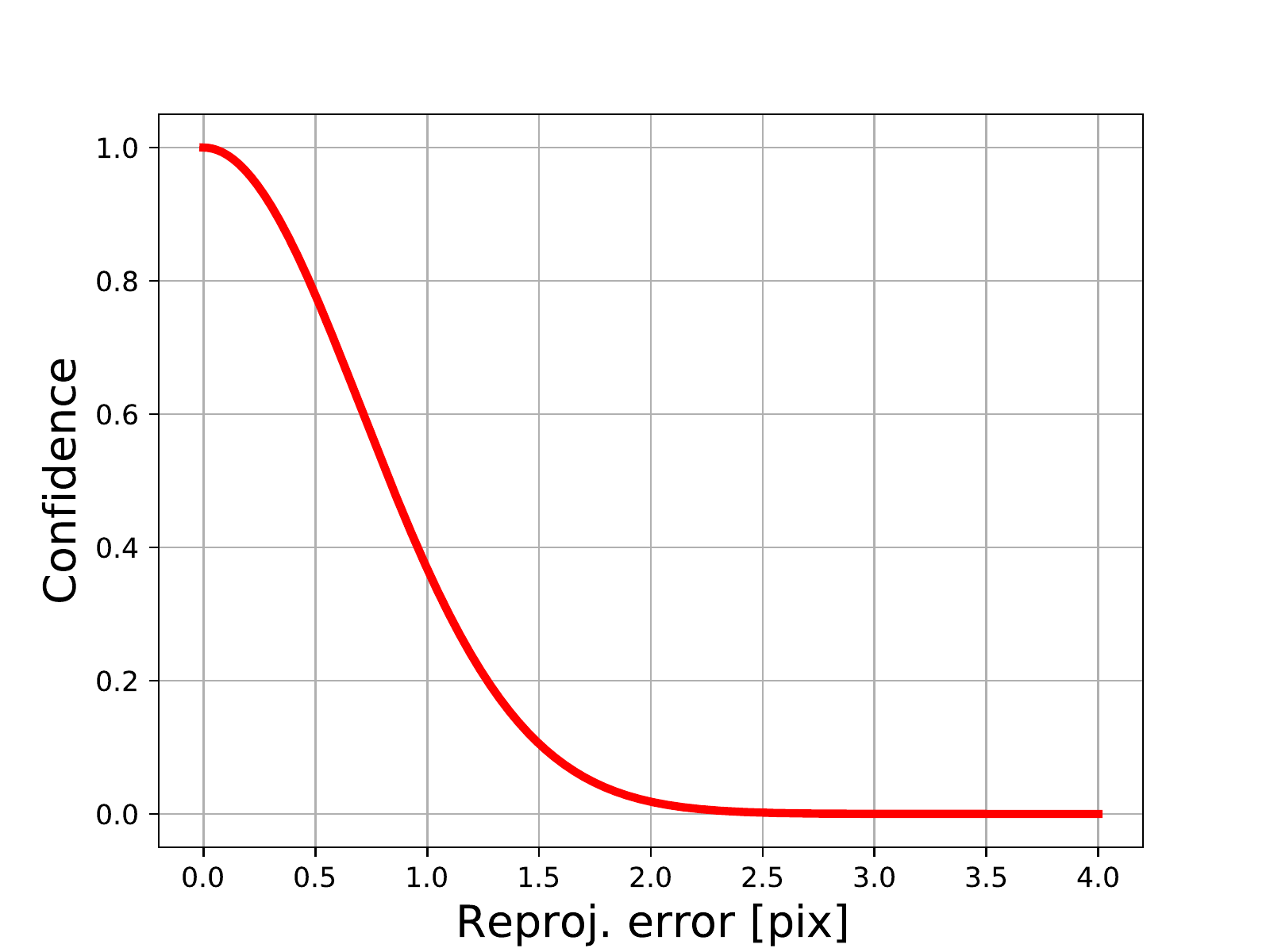}
        \caption{Confidence value as a function of the reprojection error.}
        \label{fig:confidence}
\end{figure}

The key idea of our approach is to use the result of a classical 3D reconstruction pipeline as \textit{pseudo}-ground truth for supervising the rendered geometry from NeRF. However, the MVS stage can fail in textureless areas and non-Lambertian surfaces, thus making this supervision unreliable. For this reason, we additionally estimate a pixelwise confidence for each input image. Given a pixel $(u,v)$, the corresponding confidence $c_{uv} \in [0, 1]$ is computed as:
\begin{equation}
    c_{uv} = \text{exp}\left(-\left(\frac{e_{uv}}{\Bar{e}}\right)^2\right)
\end{equation}
where $e_{uv}$ is the top-$K$ ($K = 4$) forward-backward reprojection error of pixel $(u,v)$ and $\Bar{e}$ is the mean error over the entire set of observations. This error is computed as follows. A pixel $(u_{ref},v_{ref})$ in the reference image is first projected in 3D according to its current depth estimate and observed by a source image $k$ as the pixel $(u_{src}^k,v_{src}^k)$. Then, this pixel is re-projected in 3D with the source depth hypothesis and observed again in the reference image to obtain the pixel $(\hat{u}_{ref}^k,\hat{v}_{ref}^k)$. The top-$K$ forward-backward reprojection error is simply given by:
\begin{equation}
    e_{uv} = \frac{1}{K}\sum_{k = 1}^K (u_{ref}^k - \hat{u}_{ref}^k)^2 + (v_{ref}^k - \hat{v}_{ref}^k)^2 
\end{equation}
The relationship between the confidence value and such reprojection error is shown in Fig. \ref{fig:confidence}. We have experimented with different confidence definitions, including a binary confidence with $c_{uv} = 1$ if the pixel $(u,v)$ contributed to the discrete point cloud, 0 otherwise. However, we found that softly activating each pixel with continuous confidence yields to better results.

\subsection{Loss Functions}
\label{sec:loss}

At each training iteration, a random batch of rays $\mathcal{R}$ is sampled from the dataset and differentiable volumetric rendering is used to produce both colors and geometry by integrating along the ray. Then, we propose to optimize NeRF with the sum of three losses:
\begin{equation}
    \mathcal{L} = \mathcal{L}_{rgb}(I,\hat{I}) + \lambda_{geom}\left(\mathcal{L}_{depth}(C,D,\hat{D}) + \mathcal{L}_{norm}(C,N,\hat{N})\right)
\end{equation}
Consistently with the original NeRF formulation \cite{nerf}, the first term is the standard $L_2$ loss on rendered colors, which are obtained as shown in Eq. \ref{eq:render}:
\begin{equation}
    \mathcal{L}_{rgb}(I,\hat{I}) = \sum_{\mathbf{r} \in \mathcal{R}} ||I(\mathbf{r}) - \hat{I}(\mathbf{r})||^2
\end{equation}
Moreover, we guide the optimization procedure by penalizing errors between the rendered geometry and the \textit{pseudo}-ground truth. For both depths and normals, each ray is weighted by the corresponding confidence value to softly activate the supervision only in reliable pixels. More specifically, the learned depth values are computed from Eq. \ref{eq:depthrender} and optimized as follows:
\begin{equation}
    \mathcal{L}_{depth}(C,D,\hat{D})= \sum_{\mathbf{r} \in \mathcal{R}} C(\mathbf{r}) \cdot \text{Huber}(D(\mathbf{r}), \hat{D}(\mathbf{r}))
\end{equation}
Similarly, the third loss term is the confidence-weighted loss on rendered normals, which are given by Eq. \ref{eq:normalrender}:
\begin{equation}
    \mathcal{L}_{norm}(C,N,\hat{N})= \sum_{\mathbf{r} \in \mathcal{R}} C(\mathbf{r}) \cdot \text{Huber}(N(\mathbf{r}), \hat{N}(\mathbf{r}))
\end{equation}
In both cases, the choice of the Huber loss over a standard $L_2$ loss is an additional step towards a more robust optimization. This function is quadratic for small errors and linear for large errors, making it less sensitive to outliers:

\begin{equation}
    \text{Huber}(x,y) = \begin{cases} \frac{1}{2} (x - y)^2 & \text{if} \quad |x - y| < \delta \\ \delta \cdot (|x-y| - \frac{\delta}{2}) & \text{otherwise} \end{cases}
\end{equation}

\section{Experiments}

In this section, we present experimental results to evaluate the proposed approach against the original formulation of NeRF. Software and hardware settings are detailed in Sec. \ref{sec:set}, while quantitative and qualitative results are provided in Sec. \ref{sec:quant} and Sec. \ref{sec:qual}, respectively.

\subsection{Settings}
\label{sec:set}

\paragraph{\textbf{Dataset}} We demonstrate the effectiveness of MVG-NeRF on the \textit{Truck} scene from the Tanks \& Temples dataset \cite{tnt}, as a proxy for real-world outdoor data. The scene is captured by 250 full HD images with resolution $1920 \times 1080$, as well as a high-precision laser scanner for acquiring 3D ground truth. The training set for NeRF is built by randomly selecting 90\% of the images, with the remaining 10\% being the test set. Similarly, only the training set of NeRF is used as input for classical 3D reconstruction.

\paragraph{\textbf{Software}} The calibration parameters for NeRF and the geometric \textit{pseudo}-ground truth are computed with COLMAP \cite{sfm,mvs}, a state-of-the-art classical 3D reconstruction pipeline. The results are obtained with the \textit{automatic reconstruction} mode and default parameters. NeRF optimization follows an open-source PyTorch implementation\footnote{\url{https://github.com/yenchenlin/nerf-pytorch}}, with custom modifications to support our confidence-aware geometric losses. After training, the mesh is extracted with a publicly available marching cubes algorithm\footnote{\url{https://github.com/pmneila/PyMCubes}} \cite{marching}. Confidence estimation is implemented as a C++ plugin after the MVS stage in COLMAP.

\paragraph{\textbf{Implementation Details}} The framework has been tested on a single Nvidia V100 GPU with 32 GB RAM, but it can be adapted to run on lower tier devices with less memory. NeRF is optimized for 250000 iterations and a random batch of 1024 rays is selected at each training step. For each ray, 64 and 128 points are sampled for the coarse and fine stage of hierarchical sampling, respectively. The radiance field is approximated by a 8-layer MLP with 256 neurons each, and the geometric losses are weighted with $\lambda_{geom} = 0.1$. Finally, during the mesh extraction phase, a uniform grid of $256^3$ points is fed to the marching cubes algorithm \cite{marching} and the density is thresholded at $\tau = 50$ for generating the binary occupancy field.

\subsection{Quantitative Results}
\label{sec:quant}

In this section, a numerical comparison of our approach with the basic formulation of NeRF \cite{nerf} is presented. We measure the performances of both methods in terms of the resulting 3D geometry and novel view synthesis results. Consistently with existing literature \cite{nerf,regnerf,mvsnerf}, three metrics are used to evaluate the quality of novel views:
\begin{itemize}
    \item The \textit{\textbf{P}eak \textbf{S}ignal-to-\textbf{N}oise \textbf{R}atio} (PSNR) is defined as follows:
    \begin{equation}
        \text{PSNR} = -\frac{10}{\log 10} \cdot \text{MSE}(I,\hat{I})
    \end{equation}
    where $\text{MSE}(I,\hat{I})$ is the mean squared error between the rendered and the ground truth image. Higher is better.
    \item The \textit{\textbf{S}tructural \textbf{S}imilarity \textbf{I}ndex \textbf{M}easure} (SSIM) was introduced in \cite{ssim}. It considers the perceived change in structural information, thus measuring absolute errors. Higher is better.
    \item The \textit{\textbf{L}earned \textbf{P}erceptual \textbf{I}mage \textbf{P}atch \textbf{S}imilarity} (LPIPS) was introduced in \cite{lpips}. This metric computes the similarity between the activations of two image patches in a pre-trained network, such as AlexNet \cite{alexnet}. Lower is better.
\end{itemize}

\begin{table}[t]
\centering
\setlength{\tabcolsep}{5pt}
\renewcommand{\arraystretch}{1.1}
\begin{tabular}{  l | c c c | c  }
Method & PSNR $\uparrow$ & SSIM $\uparrow$ & LPIPS $\downarrow$ & CD $\times 10^{-3}$ $\downarrow$ \\
\hline \hline
NeRF \cite{nerf} & \textbf{21.2384} & \textbf{0.6526} & \textbf{0.3819} & 2.3823 \\
NeRF w/ depth & 20.8911 & 0.6383 & 0.4318 & \underline{1.9701} \\
MVG-NeRF (ours) & \underline{21.0013} & \underline{0.6468} & \underline{0.3971} & \textbf{1.8865} \\ 
\end{tabular}
\vspace{0.2cm}
\caption{Quantitative results of NeRF with different geometric supervisions. Best and second results are \textbf{bold} and \underline{underlined}, respectively.}
\label{tab:metrics}
\end{table}

\noindent Moreover, we want to quantify the geometric results to prove that MVG-NeRF generates better 3D models. To this end, the Chamfer distance between the point cloud from the laser scanner and the mesh vertices after marching cubes \cite{marching} is computed (lower is better):
\begin{equation}
    \text{CD}(P,\hat{P}) = \frac{1}{|P|}\sum_{x \in P} \min_{y \in \hat{P}} ||x-y||^2 + \frac{1}{|\hat{P}|}\sum_{y \in \hat{P}} \min_{x \in P} ||x-y||^2
\end{equation}
The quantitative evaluation in Tab. \ref{tab:metrics} shows the comparison between our MVG-NeRF, the basic formulation of NeRF \cite{nerf} and an intermediate setting where only depth from MVS is used as supervision, without pixelwise normals. It can be seen that our approach provides a better 3D geometry, while remaining competitive on the novel view synthesis task. Moreover, note that the supervision of dense depth without normals already improves significantly the quality of the underlying scene surface. 

\subsection{Qualitative Results}
\label{sec:qual}

The significant improvement in geometry representation obtained by MVG-NeRF is also shown in qualitative results. Fig. \ref{fig:gt} visualizes the output of the classical 3D reconstruction pipeline on input images, namely the geometric priors used as \textit{pseudo}-ground truth and the confidence maps. Pixels with high confidence correspond to distinctive areas with good visibility among multiple views, while textureless regions or non-Lambertian surfaces have low confidence. This means that the loss terms on depths and normals will be softly activated only for reliable rays.

Novel RGB renderings from unseen poses are provided in Fig. \ref{fig:rgb}. Despite the differences in perceptual metrics (see Tab. \ref{tab:metrics}), the test views rendered from our MVG-NeRF match closely the output of basic NeRF. Note that the loss of quality in background regions is due to the limited sampling interval inherited from NeRF. This issue has already been solved by more recent algorithms, specifically designed for unbounded scenes \cite{nerfplusplus}, and a future research direction is to integrate such improvements within our framework (see Sec. \ref{sec:lim}).

Moreover, Fig. \ref{fig:depth} and Fig. \ref{fig:normal} show the pixelwise depths and normals obtained after volume rendering, respectively. It can be clearly seen that the proposed approach produces much smoother results, especially in terms of normal vectors. This finding is confirmed by the meshes visualized in Fig. \ref{fig:mesh}. We obtain the cleanest 3D model, without the noise of NeRF \cite{nerf} and the hallucinated geometry of COLMAP \cite{sfm,mvs}, followed by Poisson surface reconstruction \cite{poisson,screened}.

\section{Limitations and Future Work}
\label{sec:lim}

The main limitation of MVG-NeRF is that it requires the execution of a classical 3D reconstruction pipeline before training a neural radiance field, which is a significant computational burden. For this reason, our approach is suited for offline reconstruction applications, but not for real-time graphics tasks. Moreover, we identify two main research directions that will be explored as future work. 

Firstly, while we build on solid baselines for neural implicit representations \cite{nerf} and geometry-based 3D reconstruction \cite{sfm,mvs}, several improvements over both NeRF and COLMAP have been presented in recent works \cite{instant,plenoctrees,acmm,acmp}. In principle, this should accordingly improve their combination and our framework is flexible enough to support any version of these algorithms.

Secondly, beyond our novel idea of supervising NeRF with the results of a classical 3D reconstruction pipeline, another way for combining geometry and learning has been investigated in literature. Some works \cite{geoneus,neuralwarp} proposed to enforce multi-view geometry constraints by warping patches as an additional loss function. This idea could be integrated in our framework, especially for pixels with low confidence. In this context, note that the consistency between deep features patches might be more robust than simply warping raw colors.

\begin{table}
\centering
\addtolength{\tabcolsep}{1pt} 
\begin{tabular}{ c c c c}
    \multirow{1}{*}[4em]{\rotatebox{90}{Input}} &
    \includegraphics[width=0.29\linewidth]{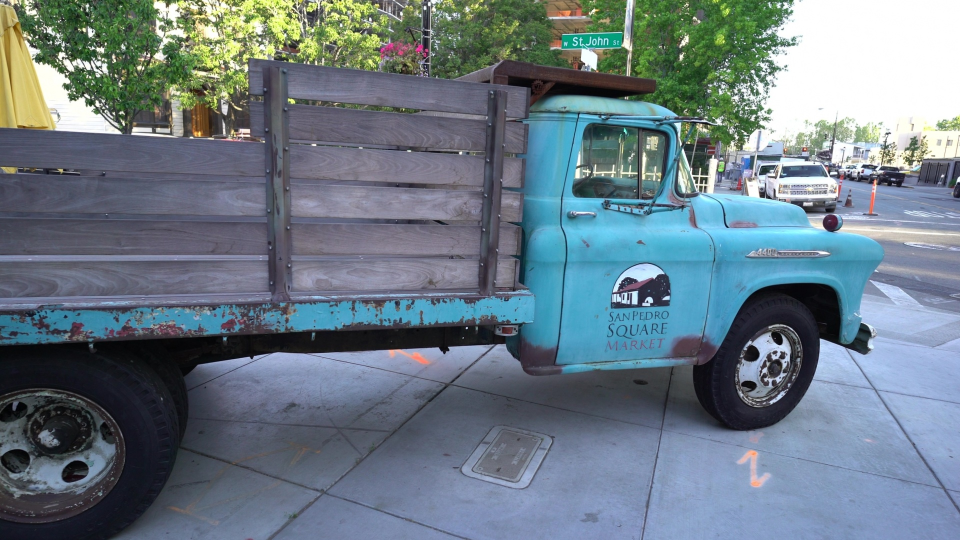} &
    \includegraphics[width=0.29\linewidth]{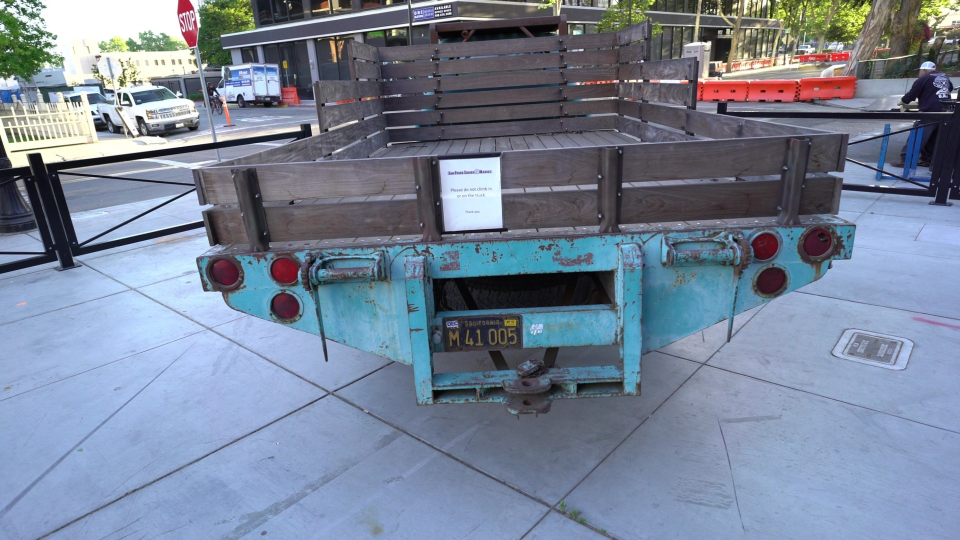} &
    \includegraphics[width=0.29\linewidth]{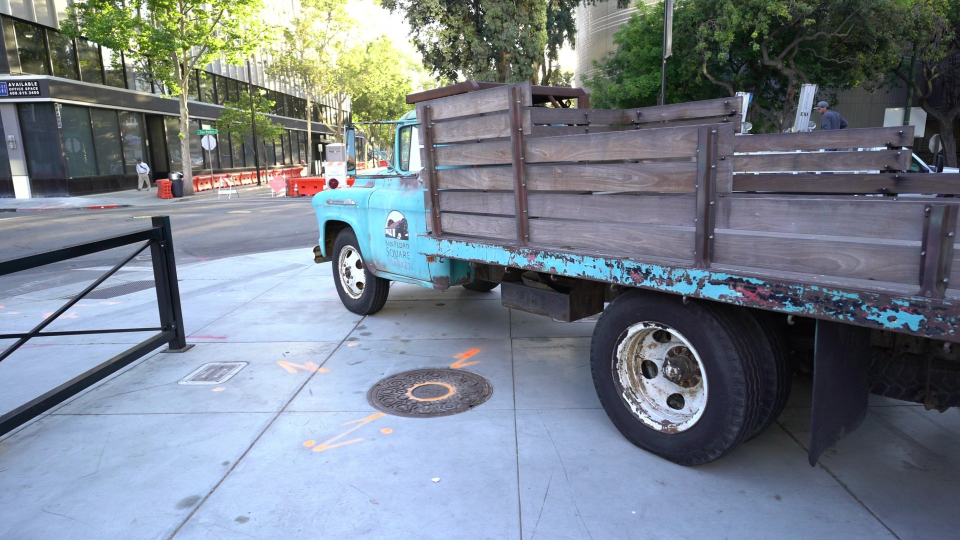} \\
    \multirow{1}{*}[4em]{\rotatebox{90}{Depth}} &
    \includegraphics[width=0.29\linewidth]{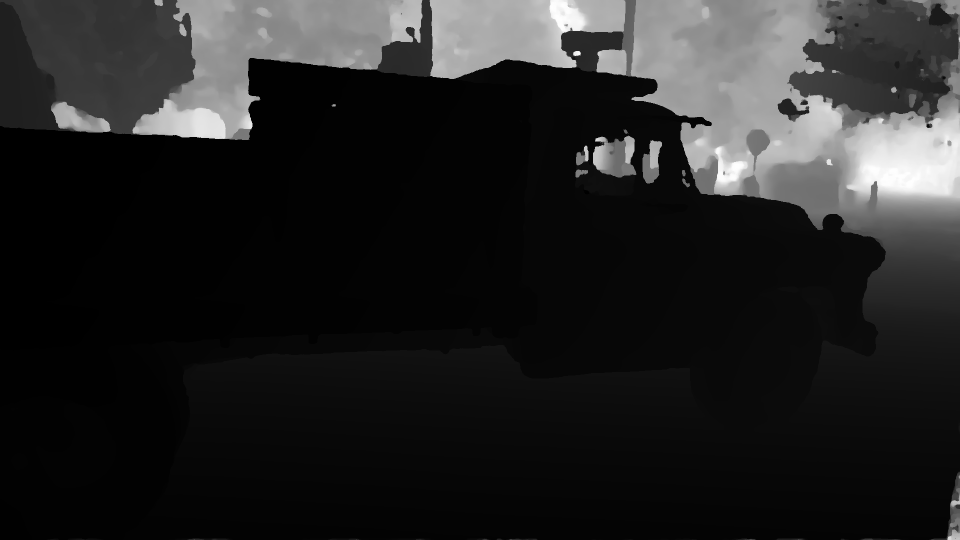} &
    \includegraphics[width=0.29\linewidth]{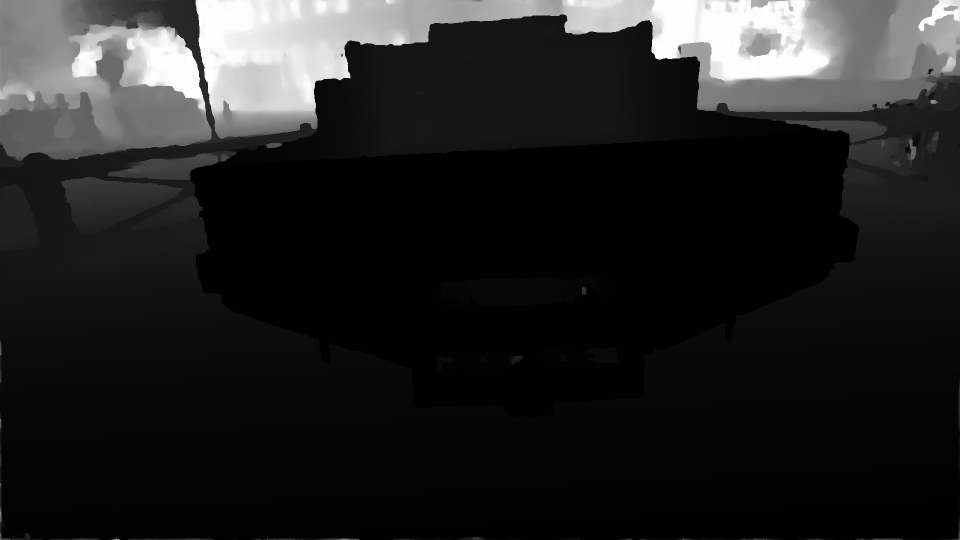} &
    \includegraphics[width=0.29\linewidth]{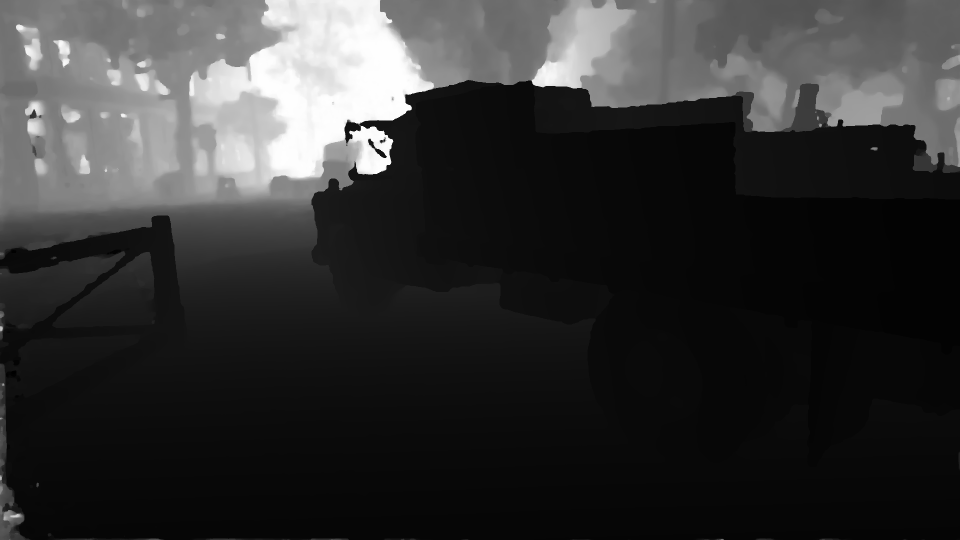} \\
    \multirow{1}{*}[4em]{\rotatebox{90}{Normal}} &
    \includegraphics[width=0.29\linewidth]{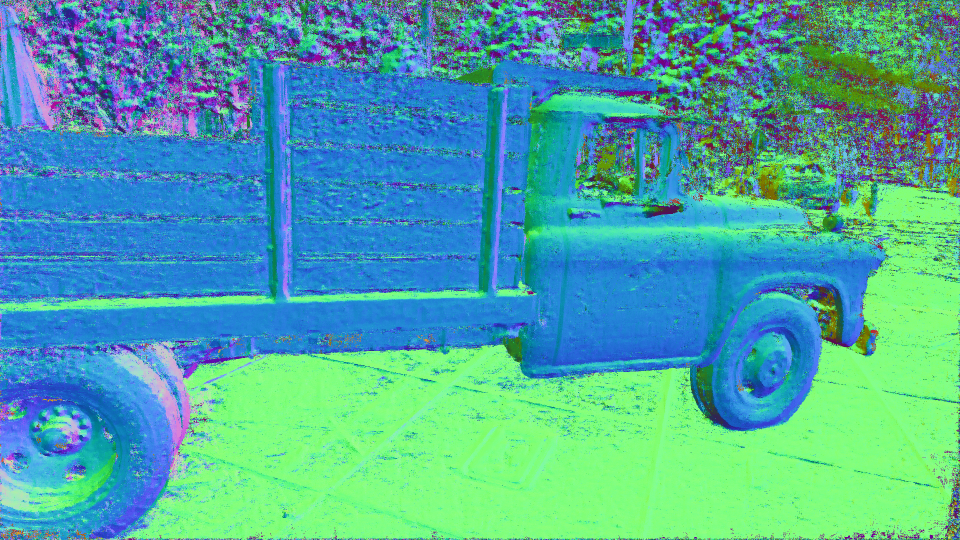} &
    \includegraphics[width=0.29\linewidth]{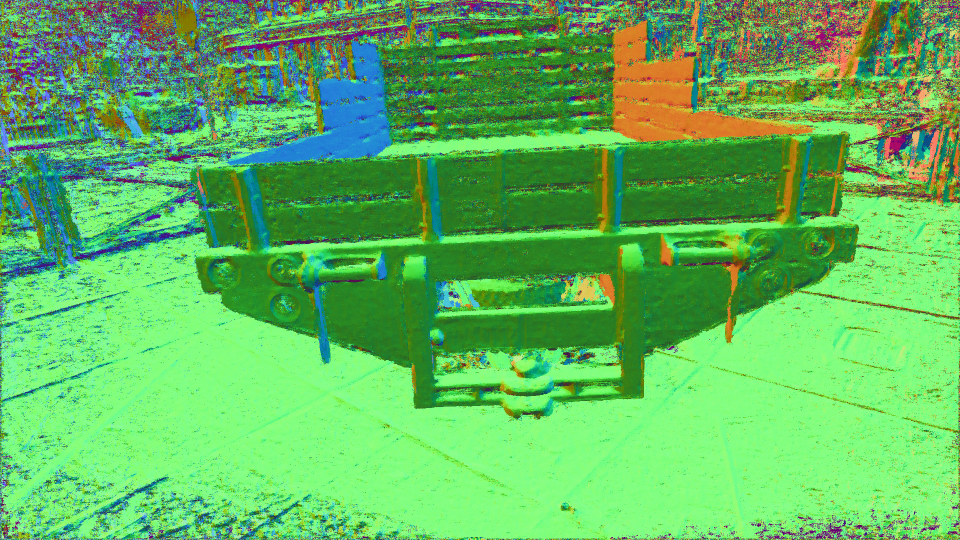} &
    \includegraphics[width=0.29\linewidth]{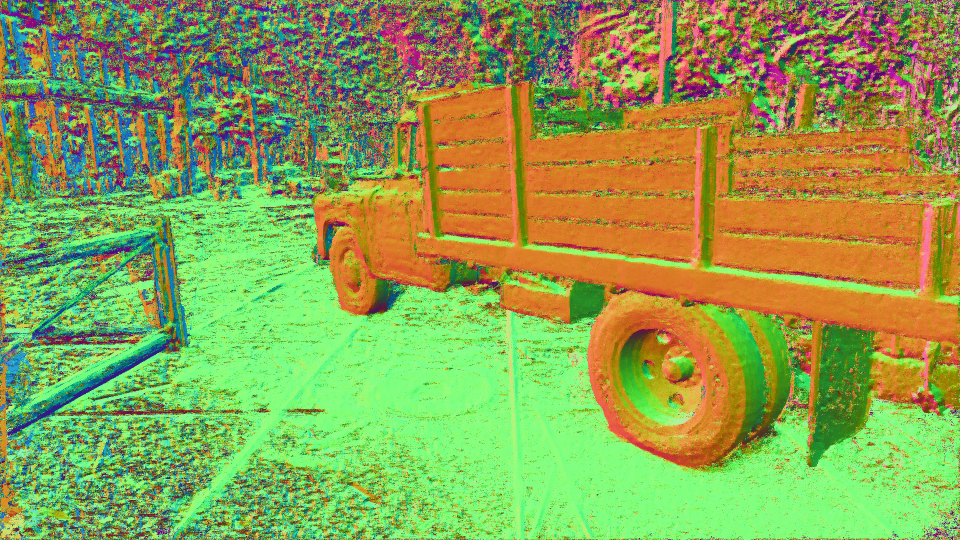} \\
    \multirow{1}{*}[5em]{\rotatebox{90}{Confidence}} &
    \includegraphics[width=0.29\linewidth]{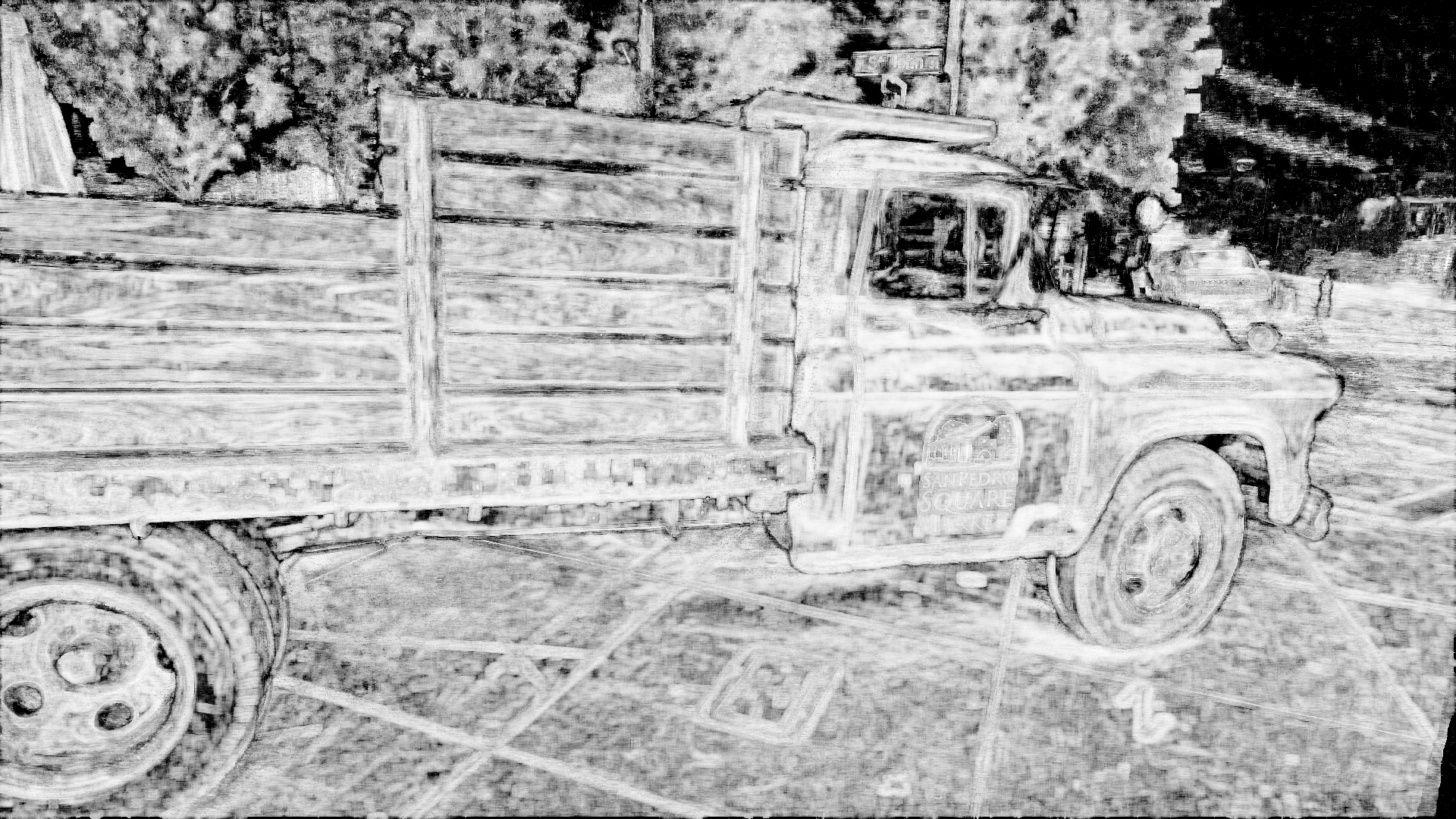} &
    \includegraphics[width=0.29\linewidth]{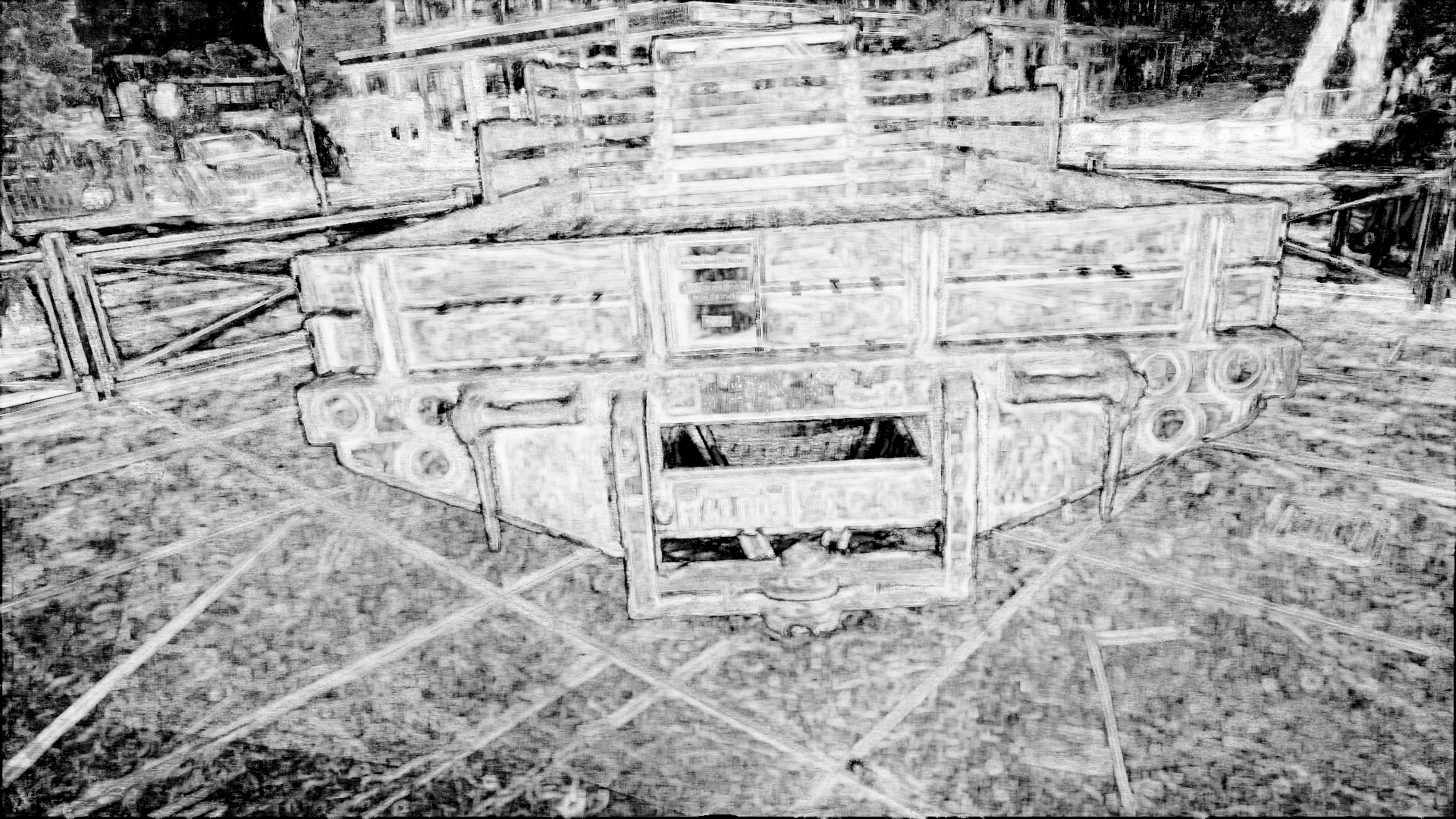} &
    \includegraphics[width=0.29\linewidth]{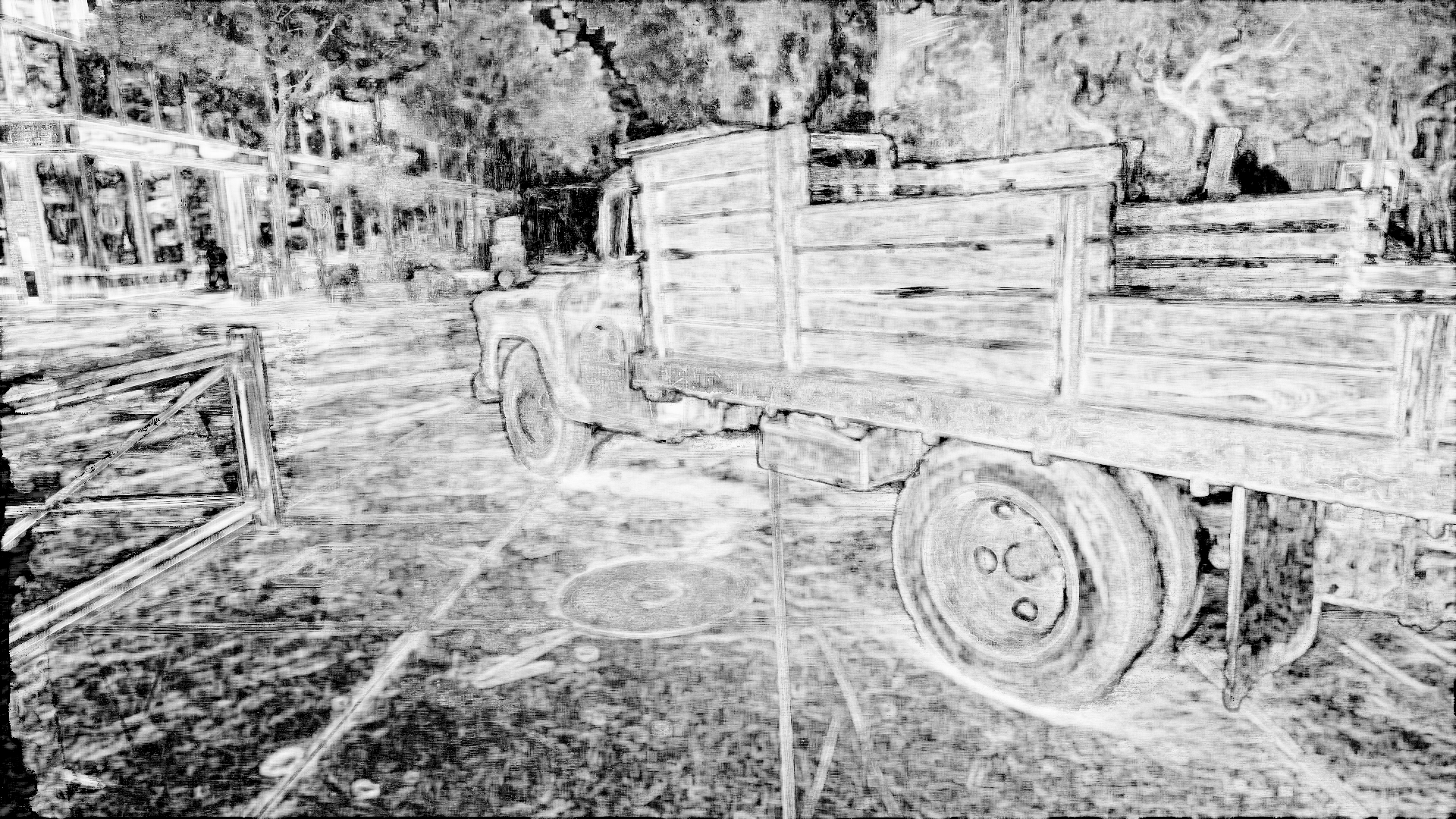} \\
\end{tabular}
\addtolength{\tabcolsep}{-1pt} 
\vspace{0.2cm}
\captionof{figure}{Geometric priors and confidence from classical 3D reconstruction \cite{sfm,mvs}.}
\label{fig:gt}
\end{table}

\begin{table}
\centering
\addtolength{\tabcolsep}{1pt} 
\begin{tabular}{ c c c c}
    \multirow{1}{*}[4em]{\rotatebox{90}{Input}} &
    \includegraphics[width=0.29\linewidth]{qual/gt/000_rgb_gt.png} &
    \includegraphics[width=0.29\linewidth]{qual/gt/014_rgb_gt.png} &
    \includegraphics[width=0.29\linewidth]{qual/gt/024_rgb_gt.png} \\
    \multirow{1}{*}[5em]{\rotatebox{90}{NeRF \cite{nerf}}} &
    \includegraphics[width=0.29\linewidth]{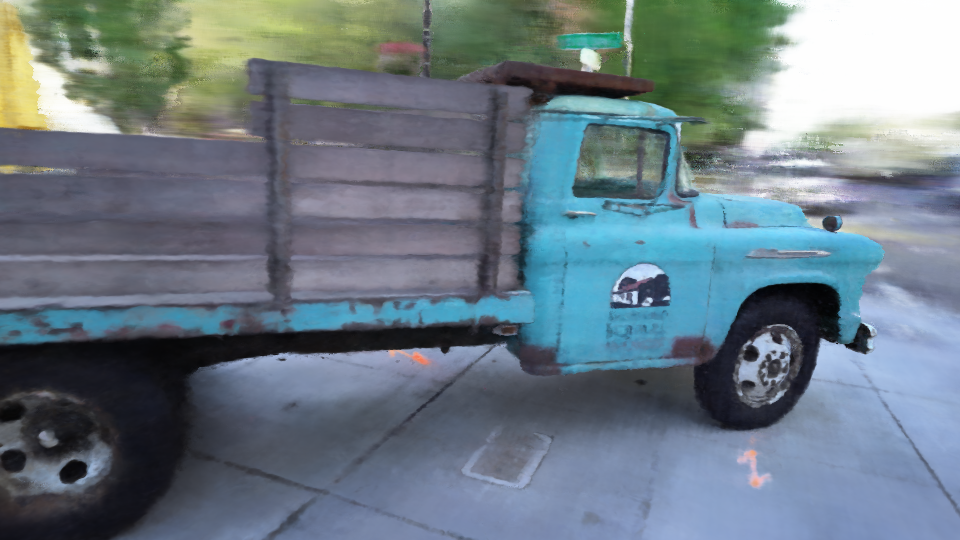} &
    \includegraphics[width=0.29\linewidth]{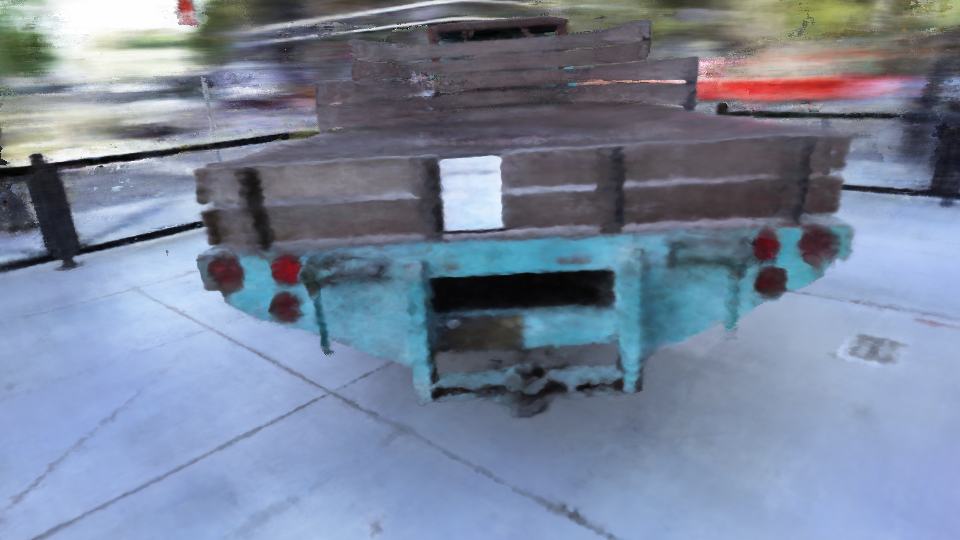} &
    \includegraphics[width=0.29\linewidth]{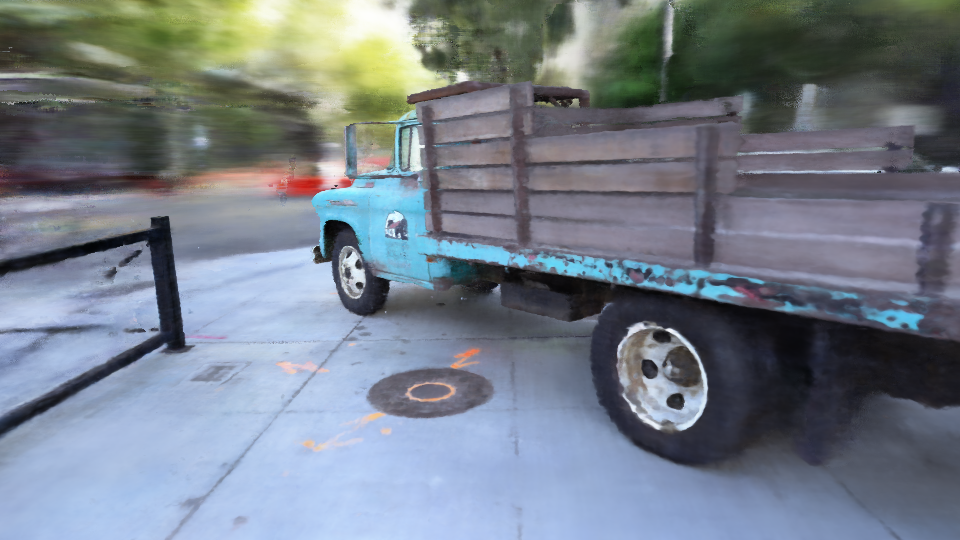} \\
    \multirow{1}{*}[5em]{\rotatebox{90}{NeRF w/ D}} &
    \includegraphics[width=0.29\linewidth]{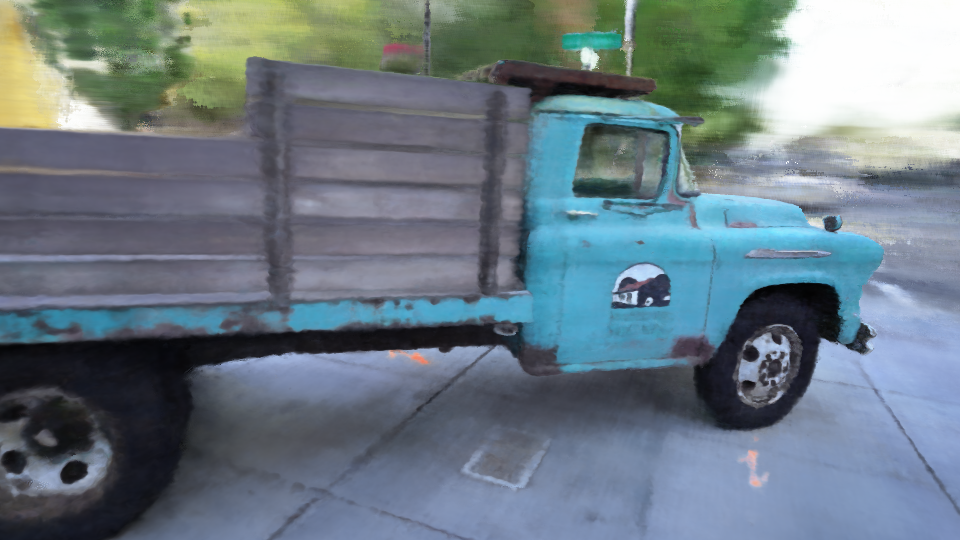} &
    \includegraphics[width=0.29\linewidth]{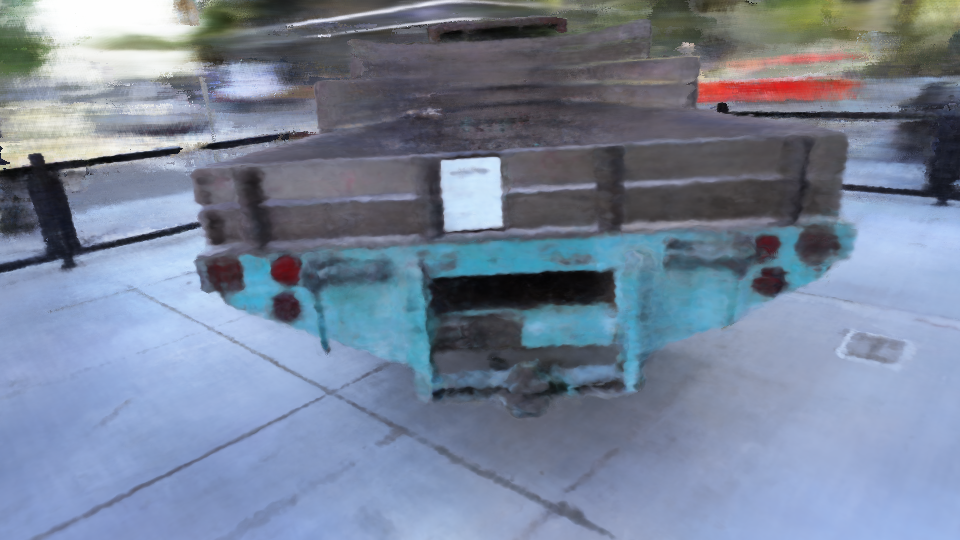} &
    \includegraphics[width=0.29\linewidth]{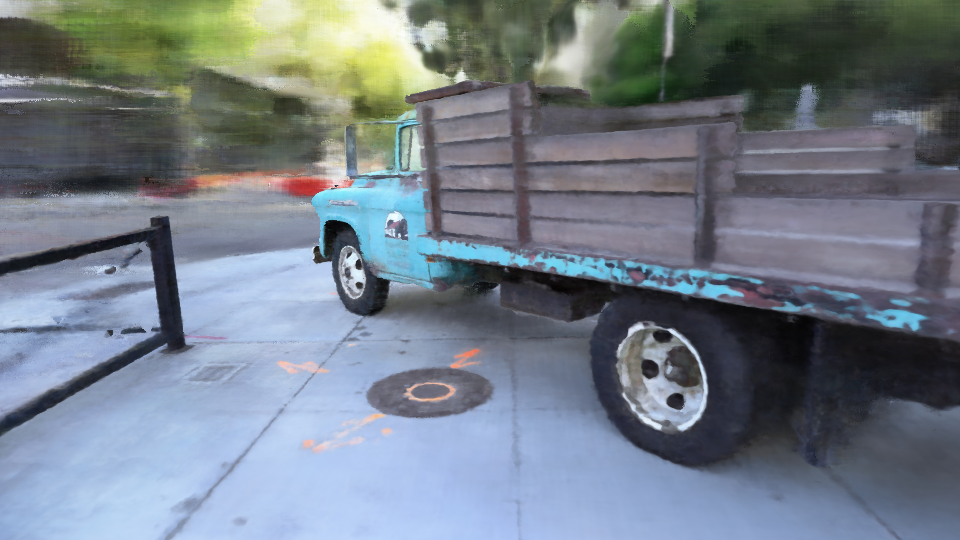} \\
    \multirow{1}{*}[5em]{\rotatebox{90}{MVG-NeRF}} &
    \includegraphics[width=0.29\linewidth]{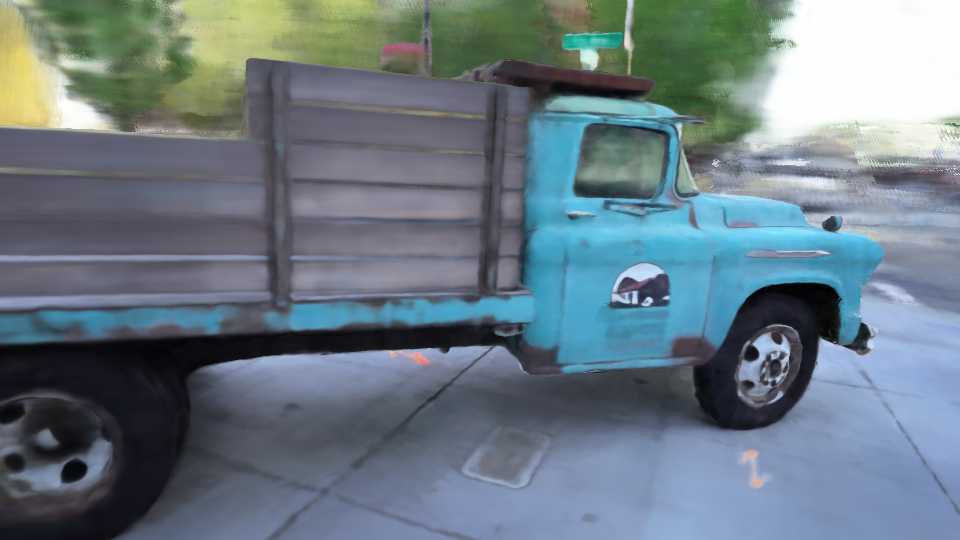} &
    \includegraphics[width=0.29\linewidth]{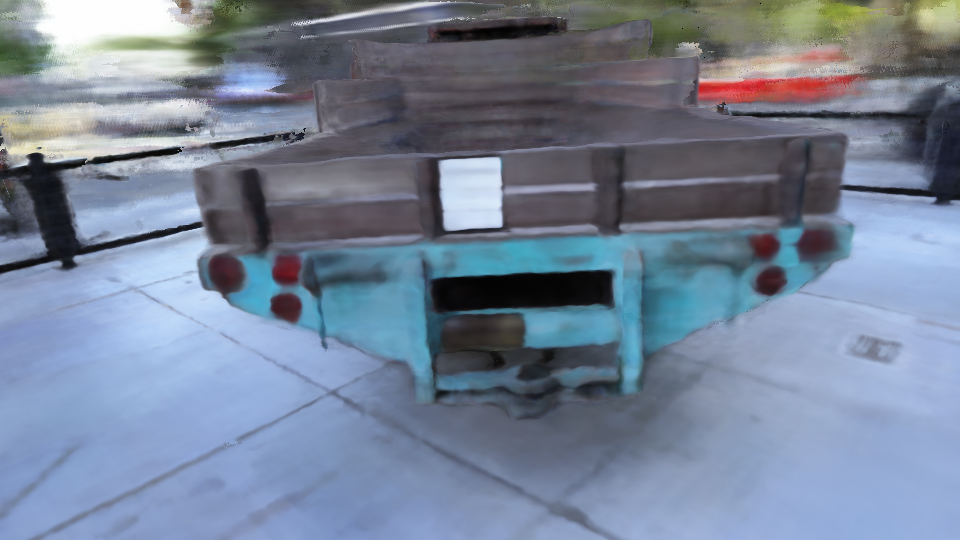} &
    \includegraphics[width=0.29\linewidth]{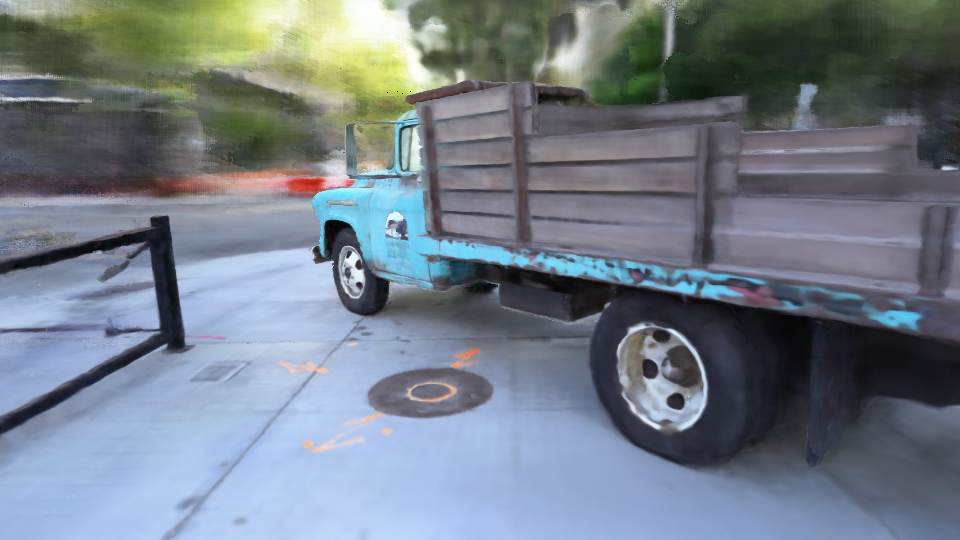} \\
\end{tabular}
\addtolength{\tabcolsep}{-1pt} 
\vspace{0.2cm}
\captionof{figure}{Qualitative comparison when rendering colors from novel views.}
\label{fig:rgb}
\end{table}

\begin{table}
\centering
\addtolength{\tabcolsep}{1pt} 
\begin{tabular}{ c c c c}
    \multirow{1}{*}[4em]{\rotatebox{90}{Input}} &
    \includegraphics[width=0.29\linewidth]{qual/gt/000_rgb_gt.png} &
    \includegraphics[width=0.29\linewidth]{qual/gt/014_rgb_gt.png} &
    \includegraphics[width=0.29\linewidth]{qual/gt/024_rgb_gt.png} \\
    \multirow{1}{*}[5em]{\rotatebox{90}{NeRF \cite{nerf}}} &
    \includegraphics[width=0.29\linewidth]{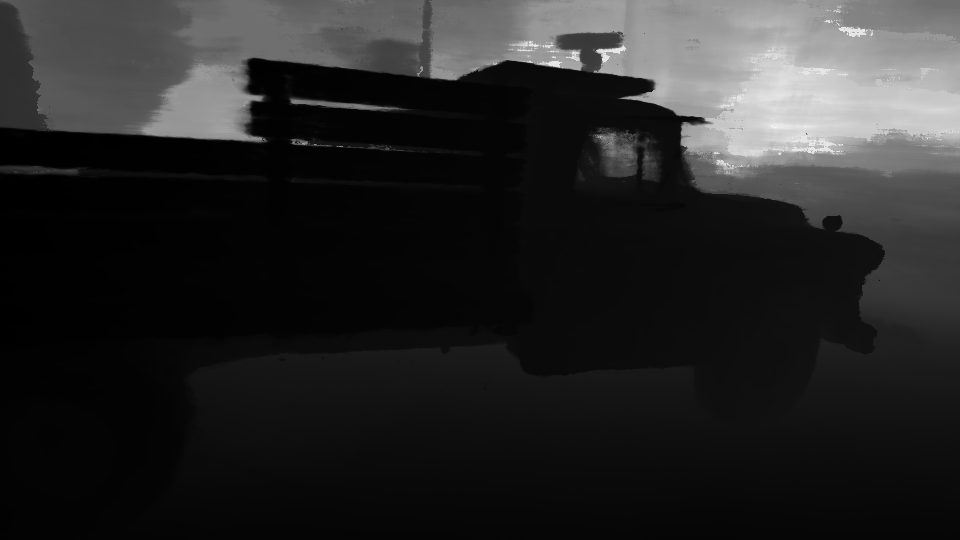} &
    \includegraphics[width=0.29\linewidth]{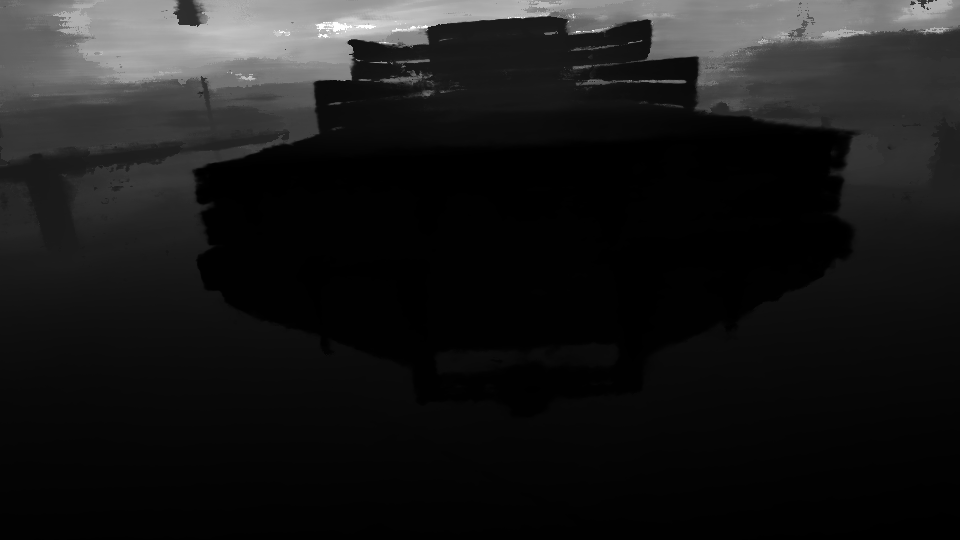} &
    \includegraphics[width=0.29\linewidth]{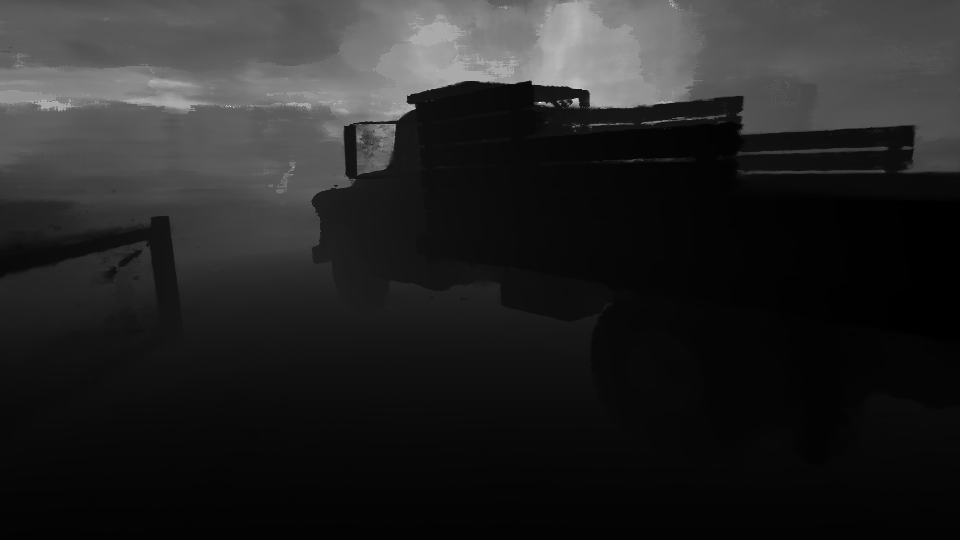} \\
    \multirow{1}{*}[5em]{\rotatebox{90}{NeRF w/ D}} &
    \includegraphics[width=0.29\linewidth]{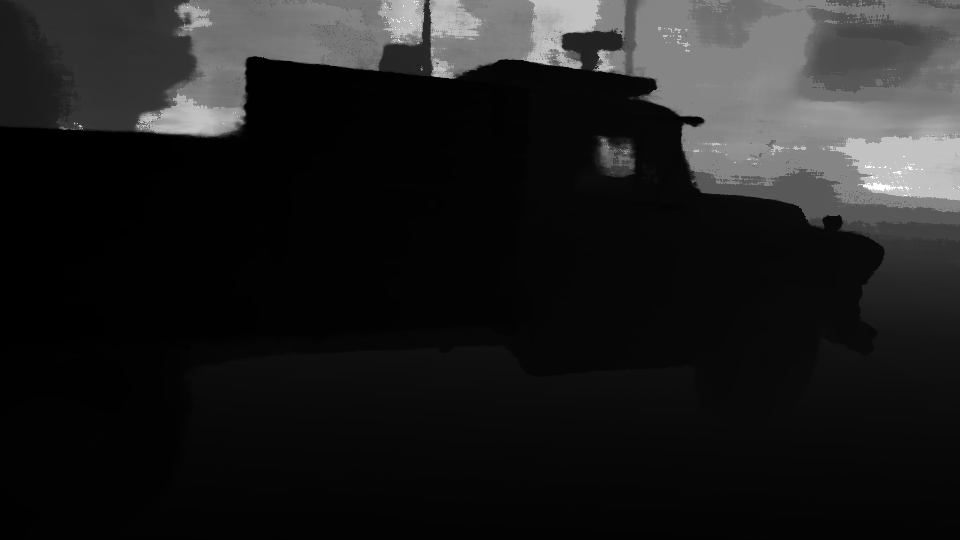} &
    \includegraphics[width=0.29\linewidth]{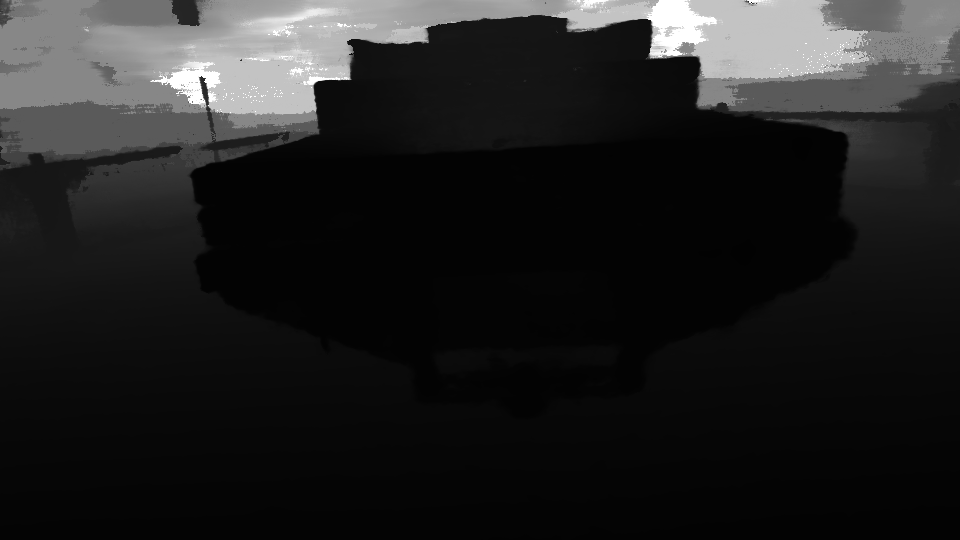} &
    \includegraphics[width=0.29\linewidth]{qual/depth/024_depth_base.png} \\
    \multirow{1}{*}[5em]{\rotatebox{90}{MVG-NeRF}} &
    \includegraphics[width=0.29\linewidth]{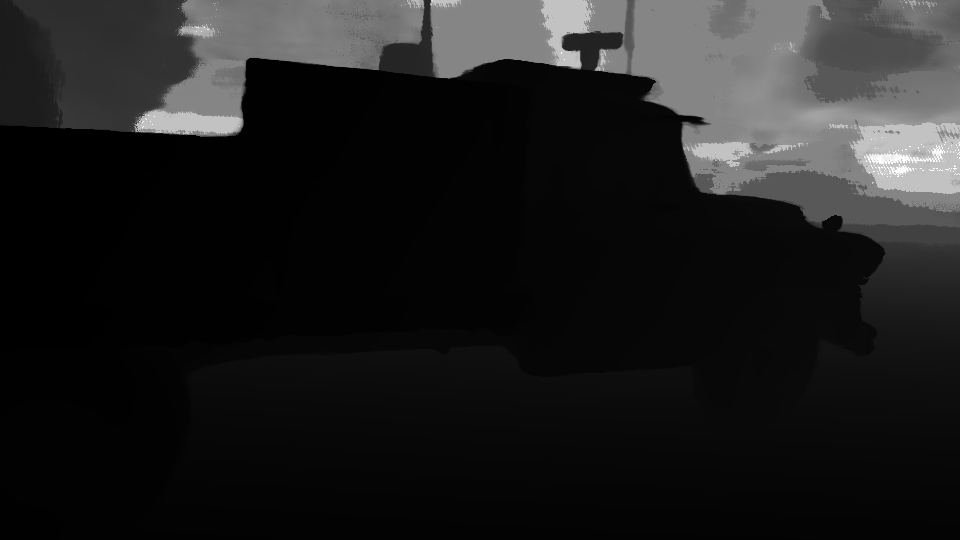} &
    \includegraphics[width=0.29\linewidth]{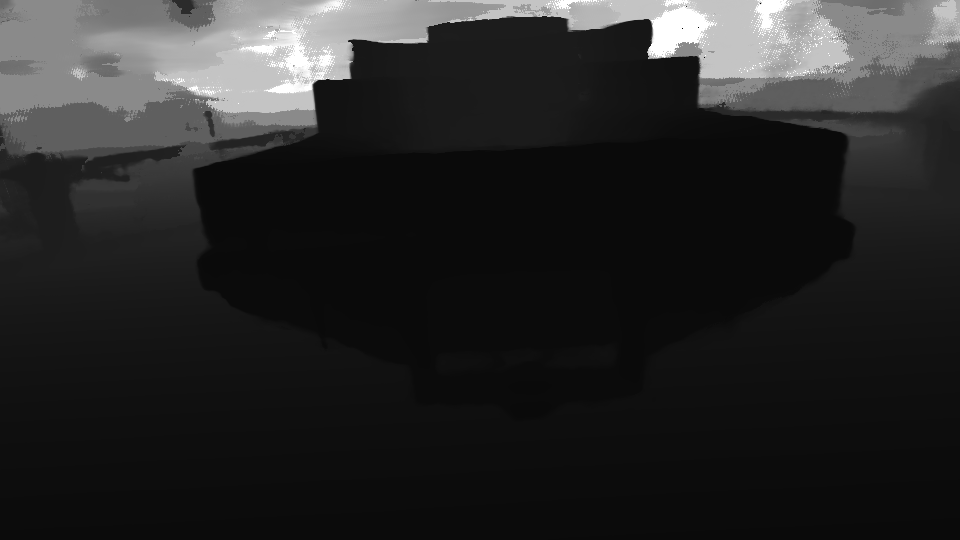} &
    \includegraphics[width=0.29\linewidth]{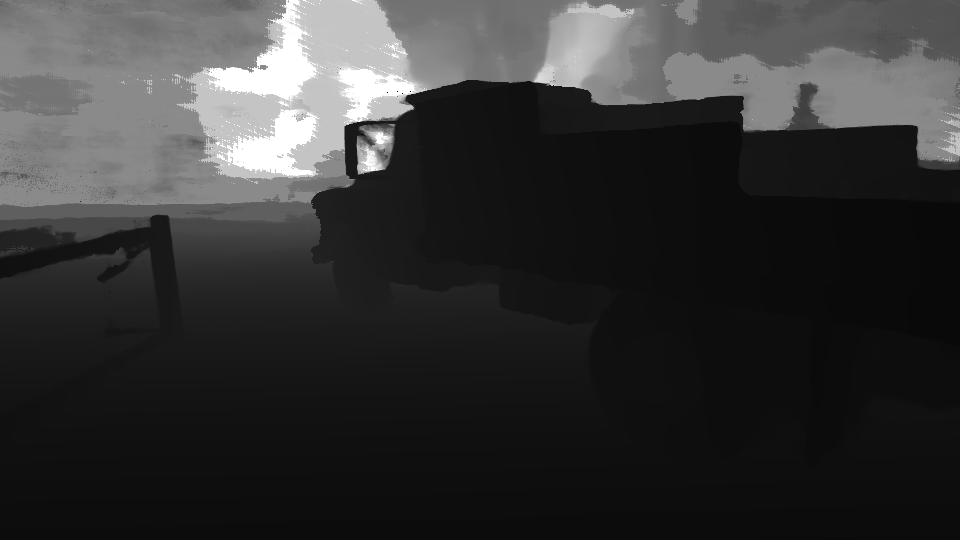} \\
\end{tabular}
\addtolength{\tabcolsep}{-1pt} 
\vspace{0.2cm}
\captionof{figure}{Qualitative comparison when rendering depths from novel views.}
\label{fig:depth}
\end{table}

\begin{table}
\centering
\addtolength{\tabcolsep}{1pt} 
\begin{tabular}{ c c c c}
    \multirow{1}{*}[4em]{\rotatebox{90}{Input}} &
    \includegraphics[width=0.29\linewidth]{qual/gt/000_rgb_gt.png} &
    \includegraphics[width=0.29\linewidth]{qual/gt/014_rgb_gt.png} &
    \includegraphics[width=0.29\linewidth]{qual/gt/024_rgb_gt.png} \\
    \multirow{1}{*}[5em]{\rotatebox{90}{NeRF \cite{nerf}}} &
    \includegraphics[width=0.29\linewidth]{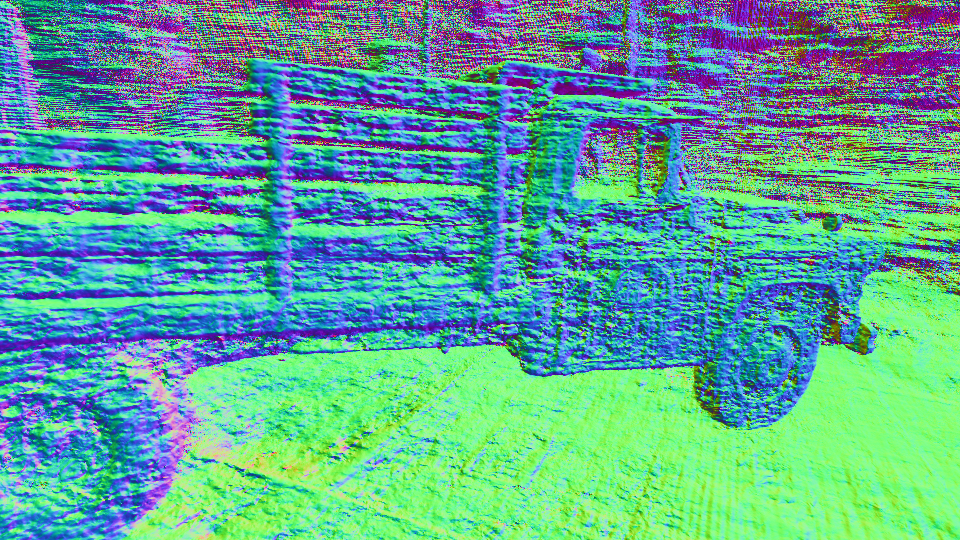} &
    \includegraphics[width=0.29\linewidth]{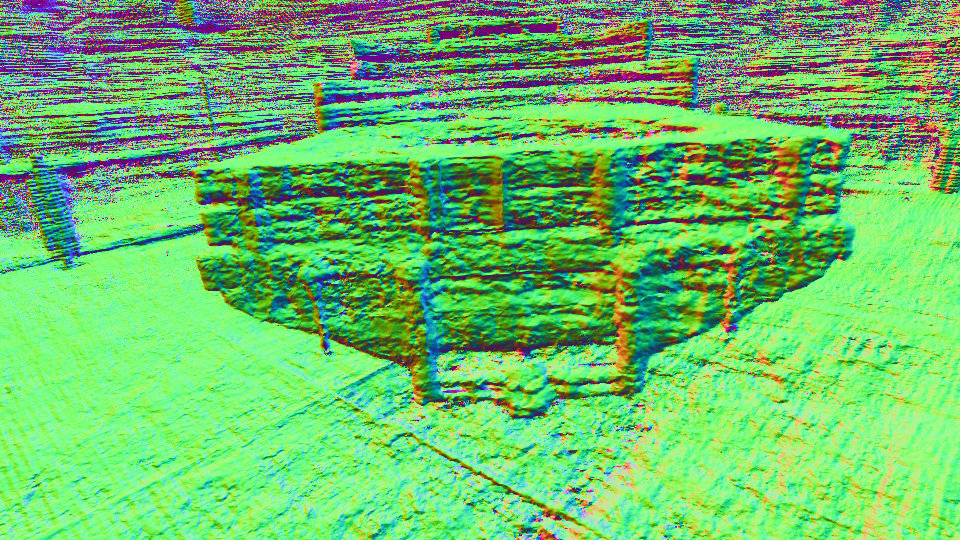} &
    \includegraphics[width=0.29\linewidth]{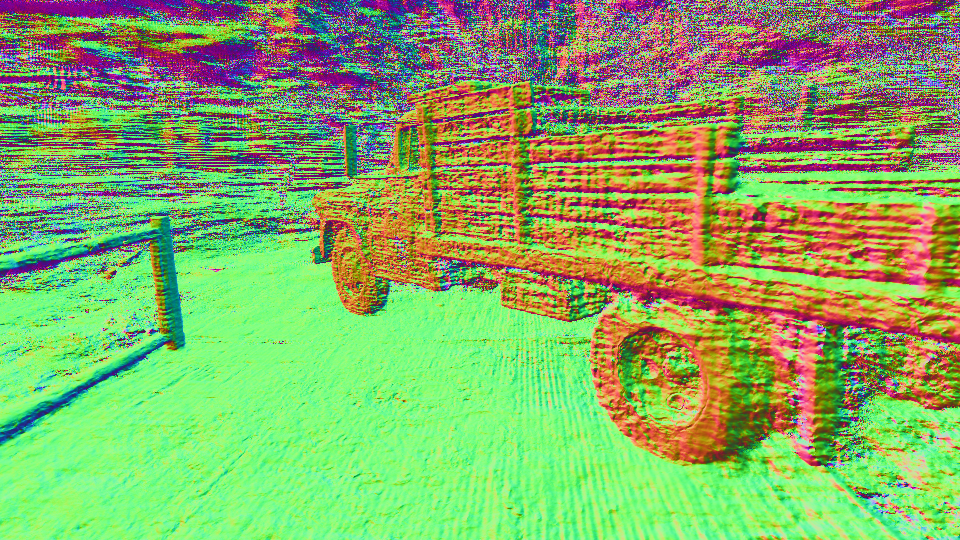} \\
    \multirow{1}{*}[5em]{\rotatebox{90}{NeRF w/ D}} &
    \includegraphics[width=0.29\linewidth]{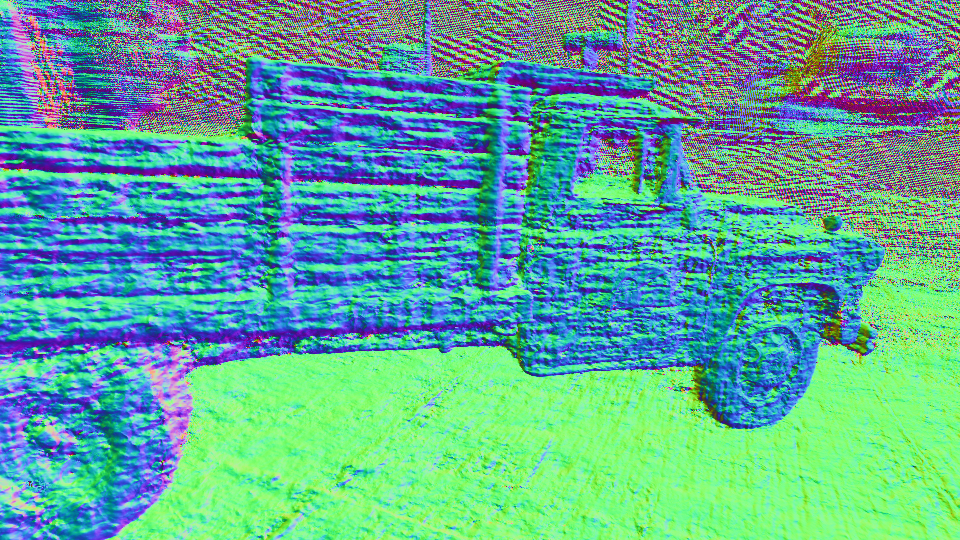} &
    \includegraphics[width=0.29\linewidth]{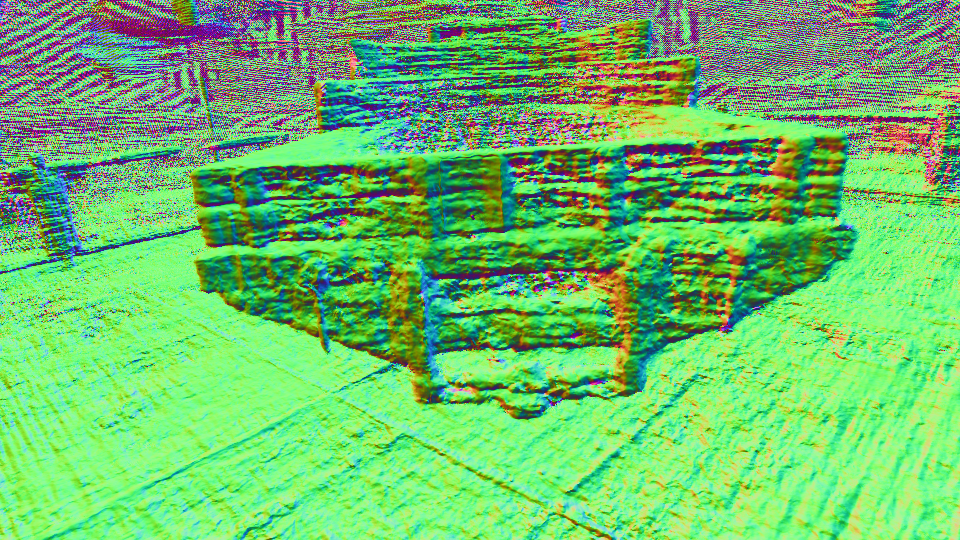} &
    \includegraphics[width=0.29\linewidth]{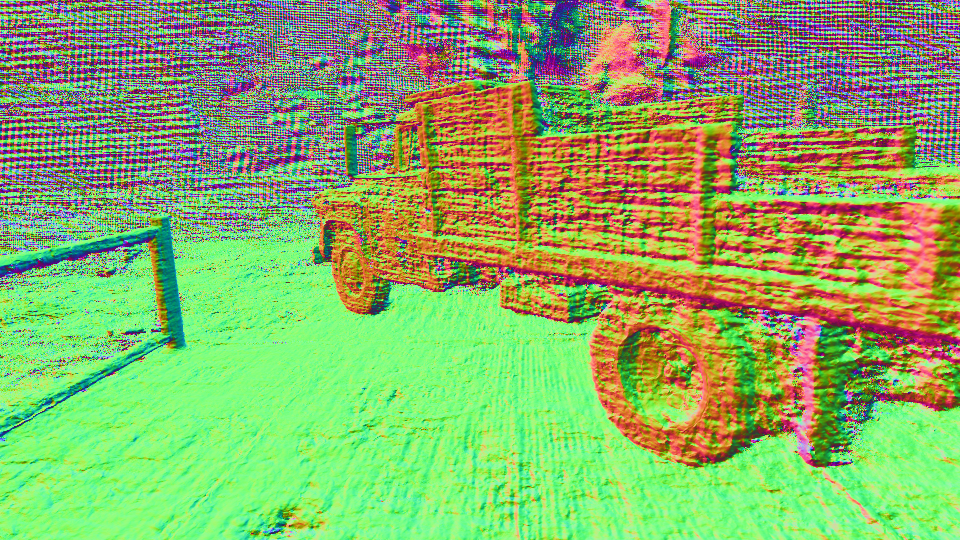} \\
    \multirow{1}{*}[5em]{\rotatebox{90}{MVG-NeRF}} &
    \includegraphics[width=0.29\linewidth]{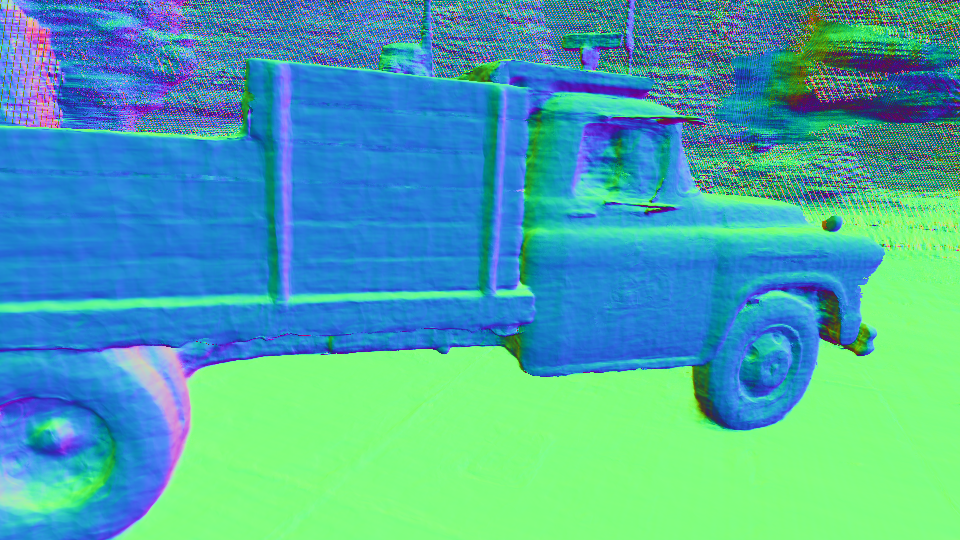} &
    \includegraphics[width=0.29\linewidth]{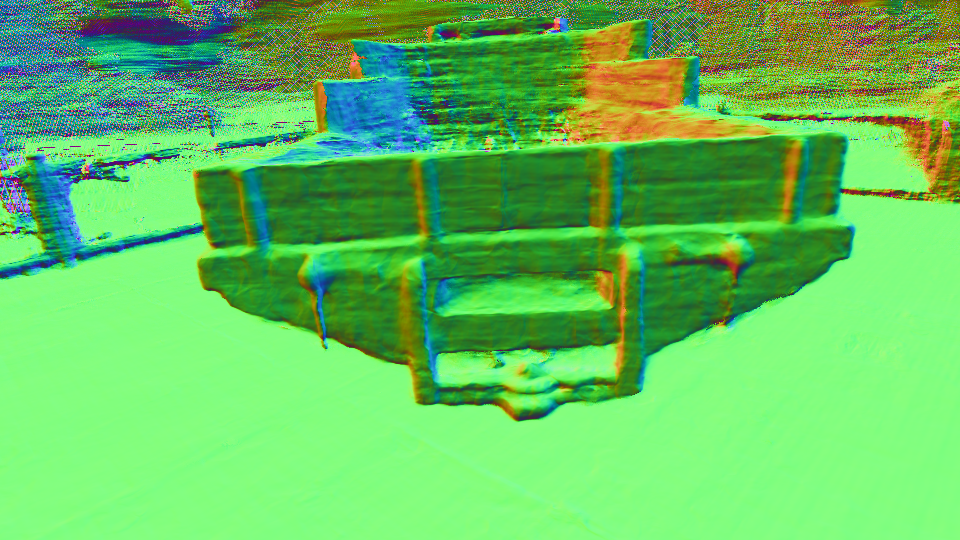} &
    \includegraphics[width=0.29\linewidth]{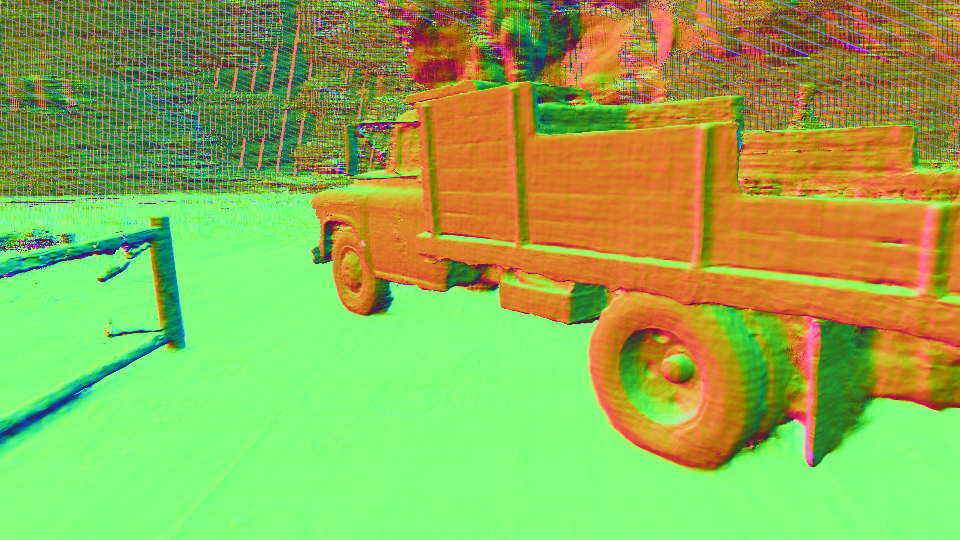} \\
\end{tabular}
\addtolength{\tabcolsep}{-1pt} 
\vspace{0.2cm}
\captionof{figure}{Qualitative comparison when rendering normals from novel views.}
\label{fig:normal}
\end{table}

\section{Conclusion}

In this paper, we propose MVG-NeRF, a framework that effectively supervise NeRF geometry with classical 3D reconstruction during training, in order to generate cleaner and smoother 3D shapes. The key idea is to compute poses and calibration parameters with SFM, as well as pixelwise depths and normals for each input view with a state-of-the-art MVS algorithm. These geometric priors are used as \textit{pseudo}-ground truth to guide NeRF optimization towards a multi-view consistent solution. Moreover, confidence maps are estimated to softly activated such supervision only in reliable pixels with low reprojection error. In this way, the confidence-aware geometric losses ignore the \textit{pseudo}-ground truth in textureless areas and non-Lambertian surface, where MVS algorithms are known to fail. We show that MVG-NeRF significantly improves the resulting mesh quality on real-world outdoor scenes, while maintaining competitive performances on the novel view synthesis task.

\begin{table}[t]
\centering
\addtolength{\tabcolsep}{1pt}
\begin{tabular}{ c c c c}
    \multirow{1}{*}[4em]{\rotatebox{90}{COLMAP}} &
    \includegraphics[width=0.33\linewidth]{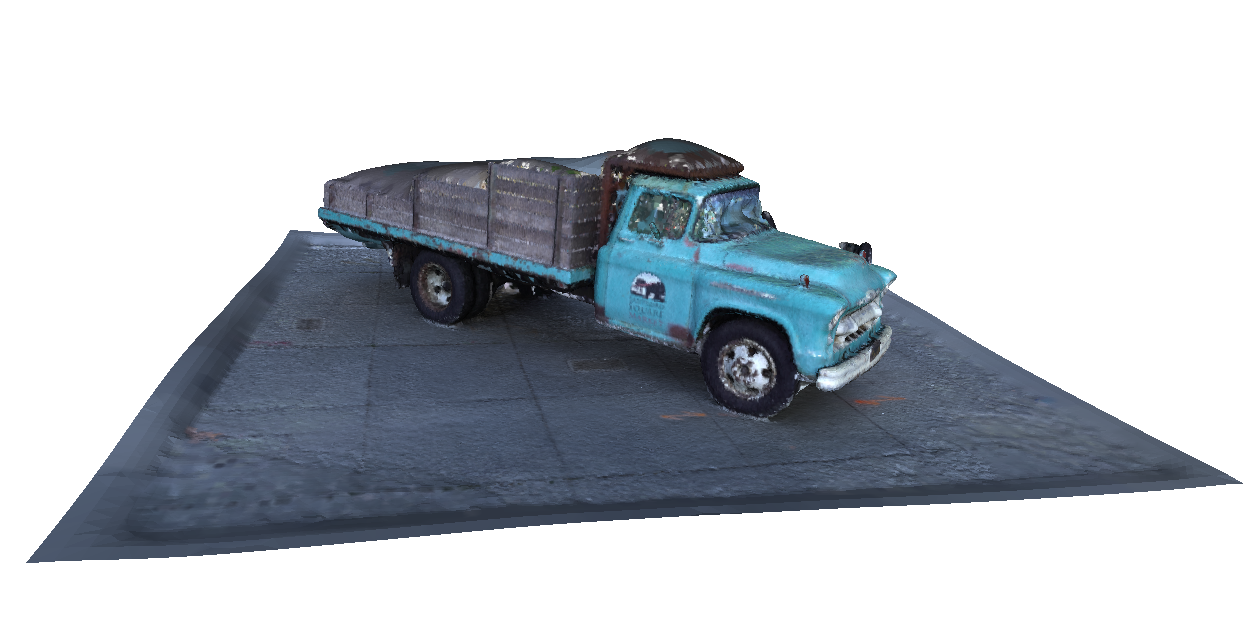} &
    \includegraphics[width=0.33\linewidth]{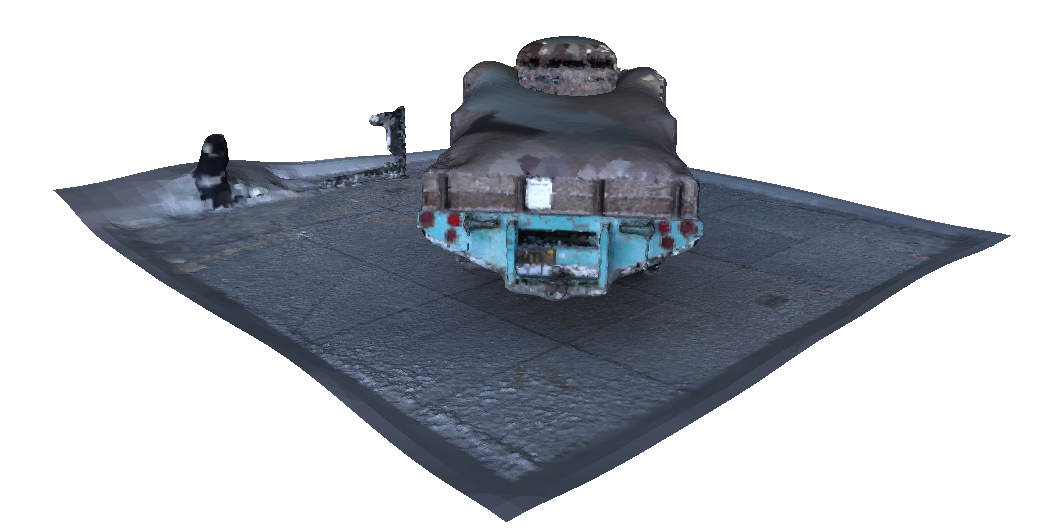} &
    \includegraphics[width=0.33\linewidth]{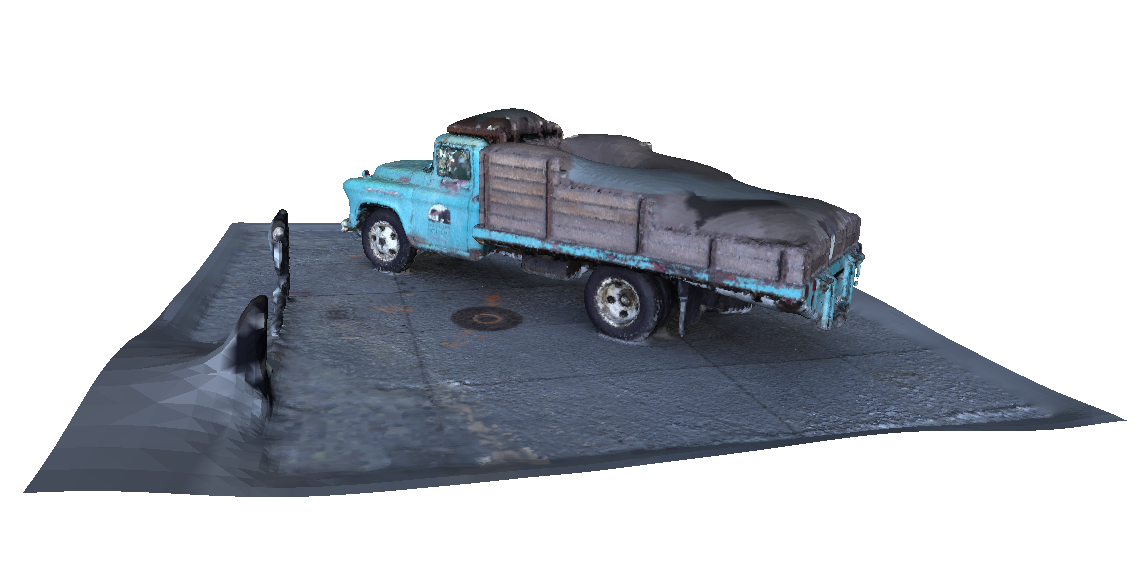} \\
    \multirow{1}{*}[5em]{\rotatebox{90}{NeRF \cite{nerf}}} &
    \includegraphics[width=0.33\linewidth]{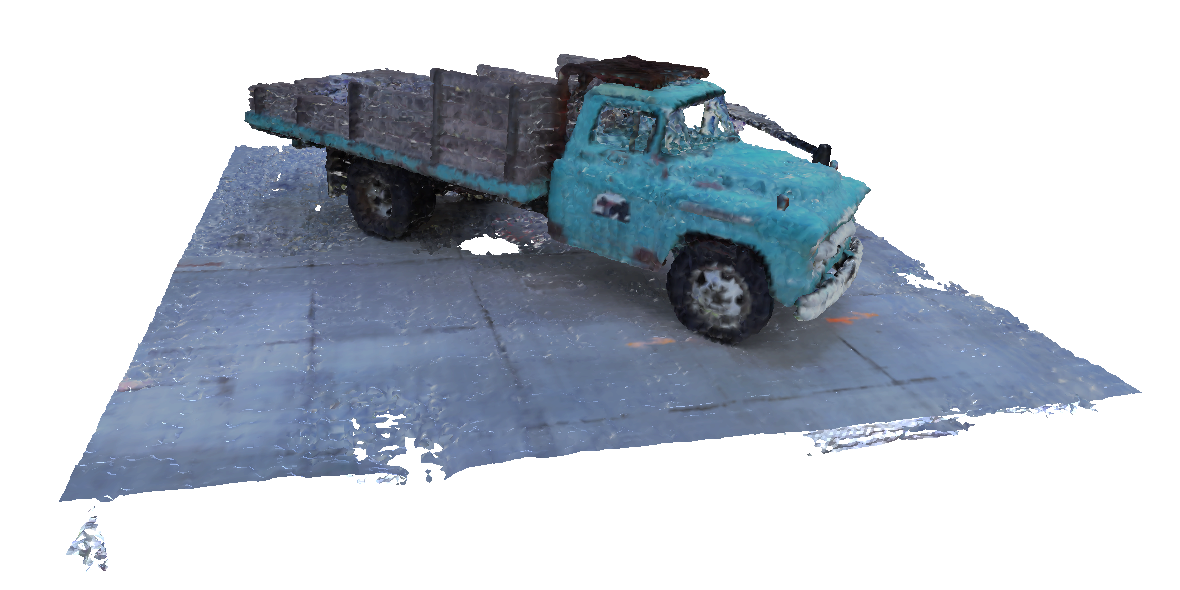} &
    \includegraphics[width=0.33\linewidth]{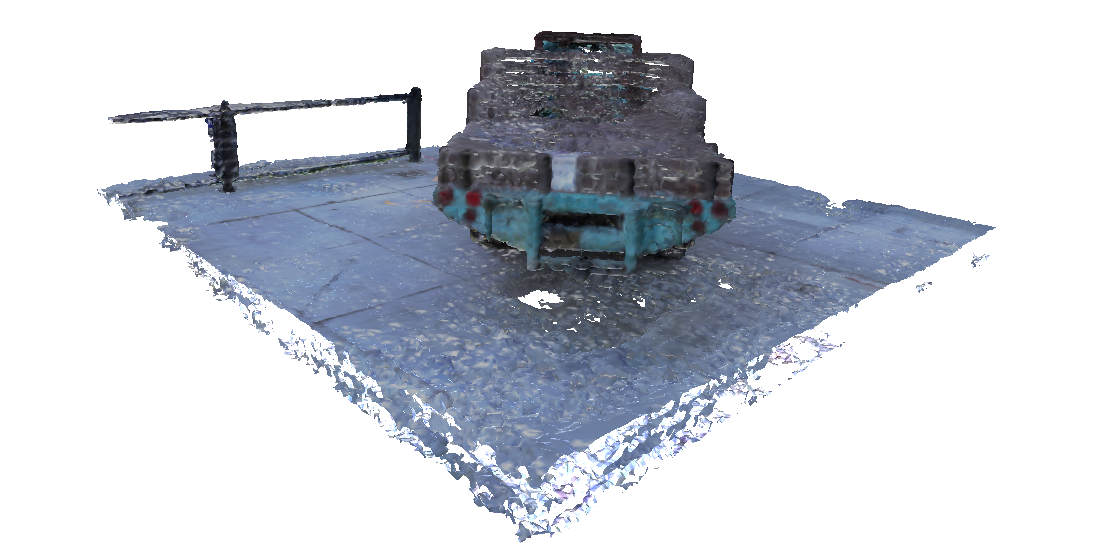} &
    \includegraphics[width=0.33\linewidth]{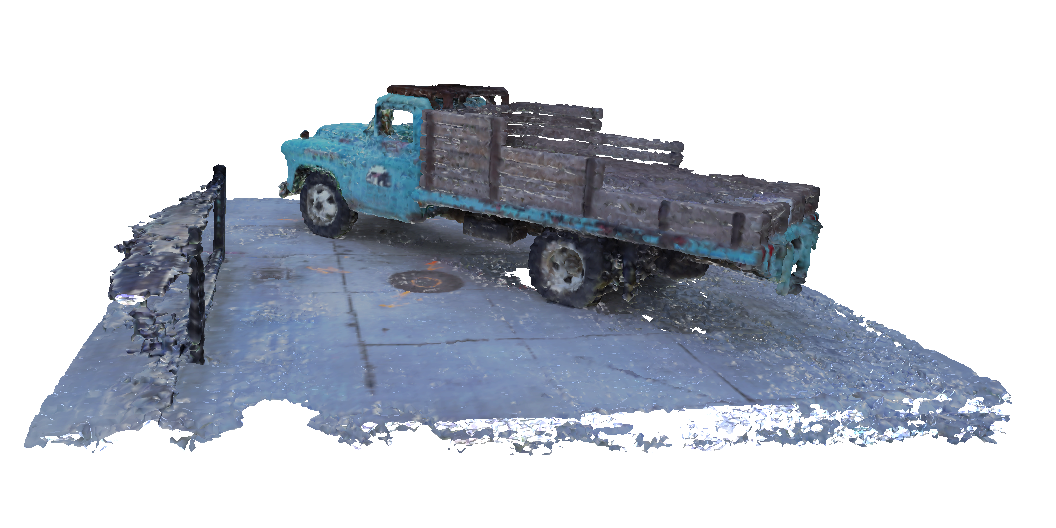} \\
    \multirow{1}{*}[5em]{\rotatebox{90}{NeRF w/ D}} &
    \includegraphics[width=0.33\linewidth]{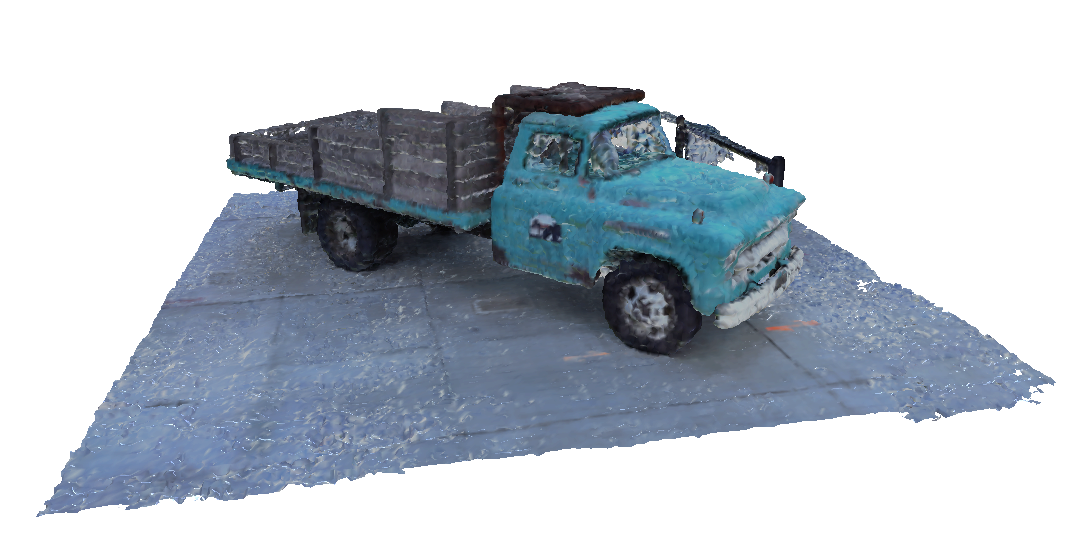} &
    \includegraphics[width=0.33\linewidth]{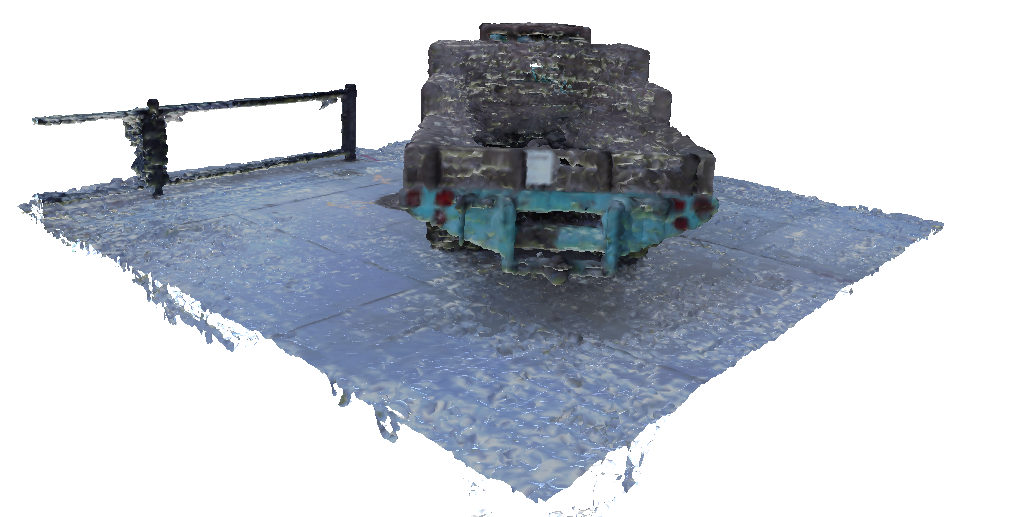} &
    \includegraphics[width=0.33\linewidth]{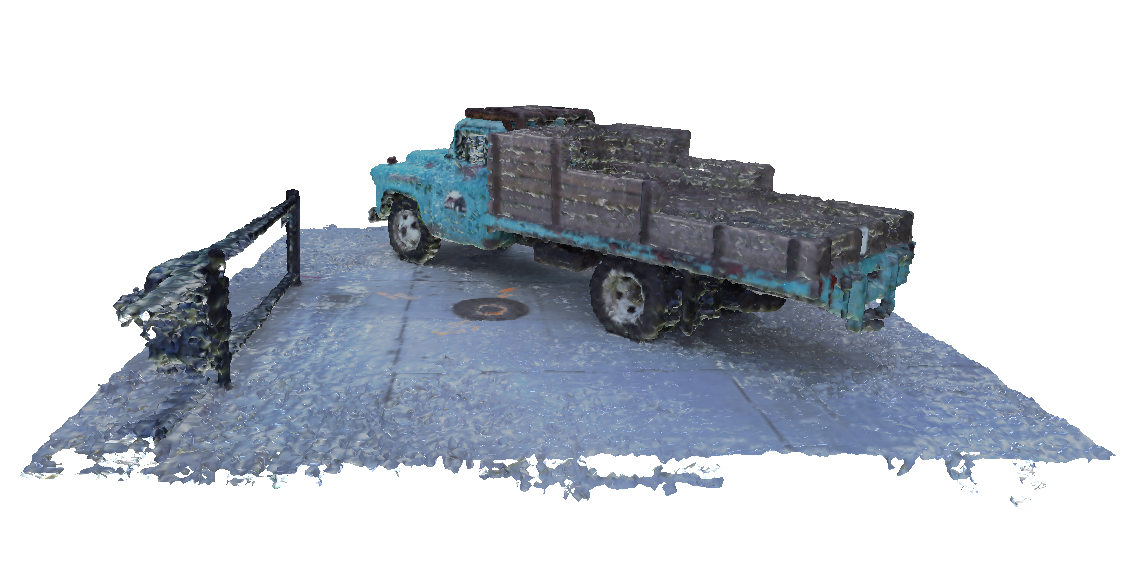} \\
    \multirow{1}{*}[5em]{\rotatebox{90}{MVG-NeRF}} &
    \includegraphics[width=0.33\linewidth]{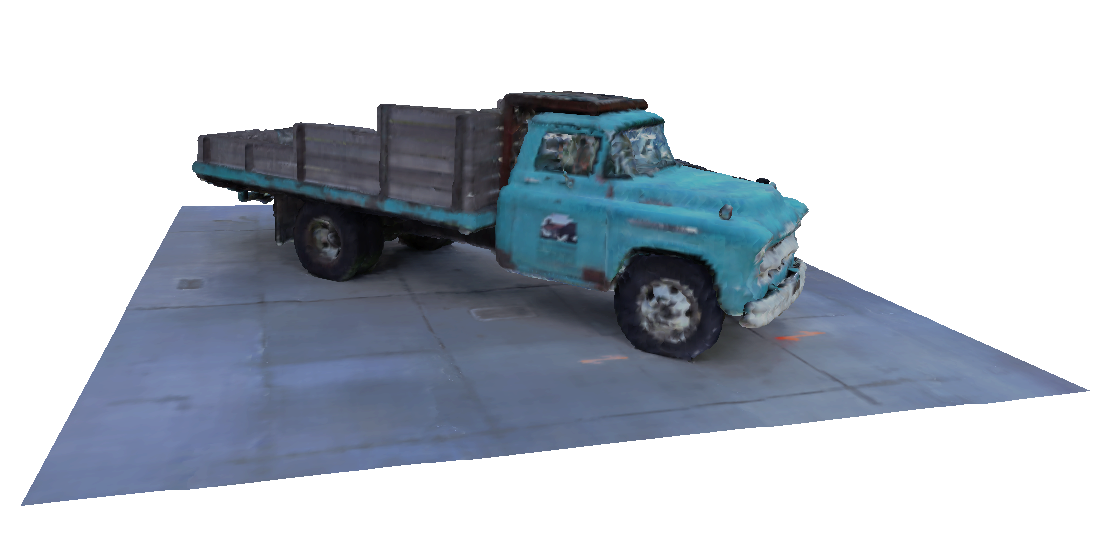} &
    \includegraphics[width=0.33\linewidth]{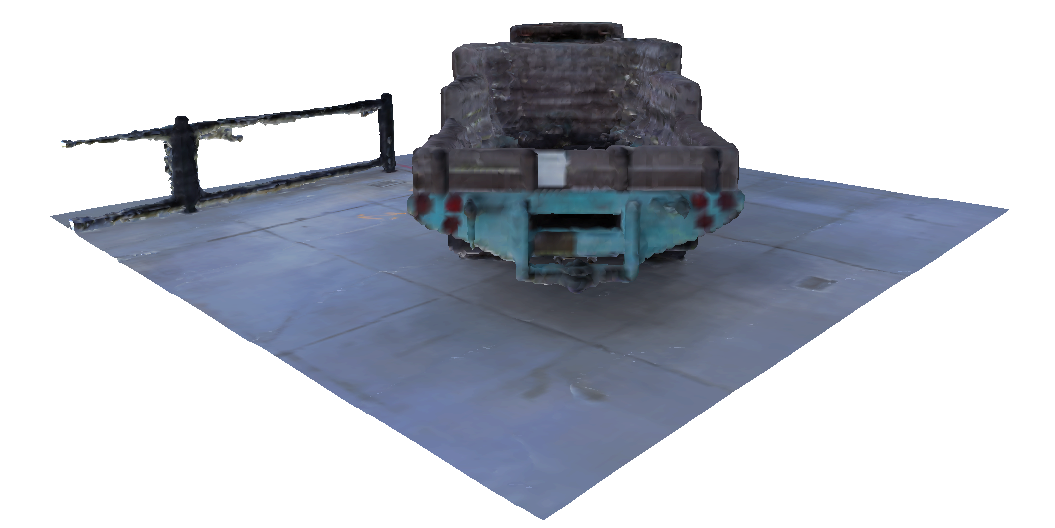} &
    \includegraphics[width=0.33\linewidth]{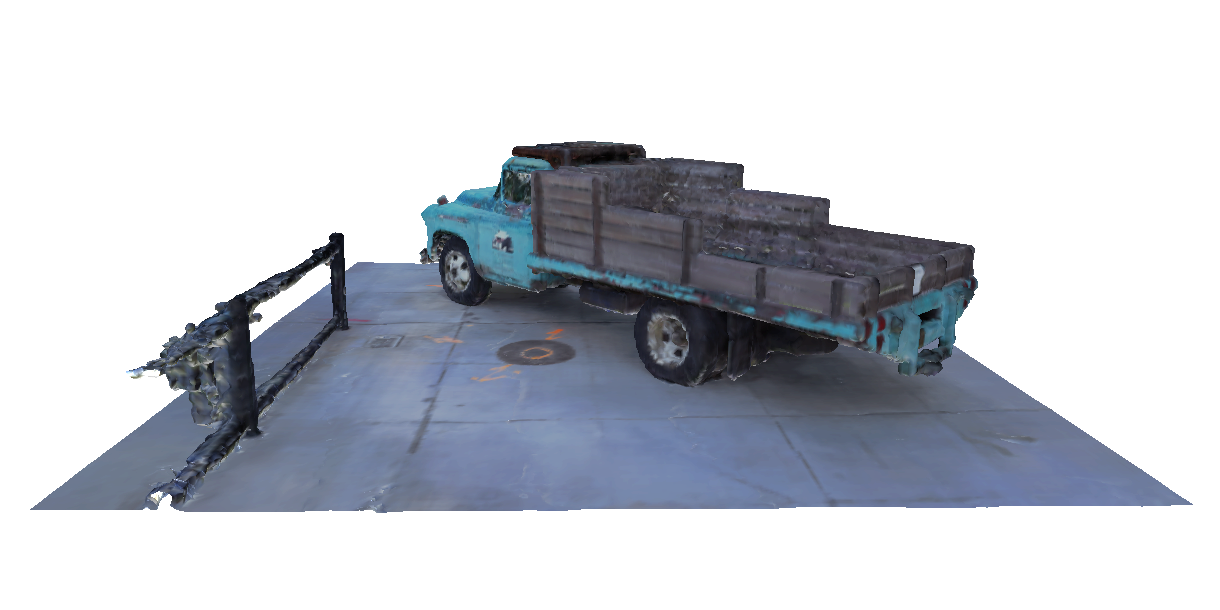} \\
\end{tabular}
\addtolength{\tabcolsep}{-1pt} 
\vspace{0.2cm}
\captionof{figure}{Qualitative comparison of the resulting 3D mesh. The first row shows the result of Poisson surface reconstruction \cite{poisson,screened}, applied on the dense point cloud from COLMAP \cite{sfm,mvs}. Our approach removes both the noise of NeRF and the hallucinated geometry of the classical pipeline.}
\label{fig:mesh}
\end{table}

\clearpage
%
%
\bibliographystyle{splncs04}
\bibliography{egbib}
\end{document}